\documentclass[11pt]{article}

\usepackage{acl}
\usepackage{chngcntr} 
\usepackage{graphicx}
\usepackage{times}
\usepackage{latexsym}
\usepackage{hyperref}
\usepackage{amsmath}
\usepackage{booktabs}
\usepackage{stfloats}
\usepackage{multirow}
\usepackage{tabularx}
\usepackage{graphicx}
\usepackage{hyperref}
\usepackage{cleveref}
\usepackage{float}
\usepackage[T1]{fontenc}

\usepackage{algorithm}
\usepackage{algorithmic}
\usepackage{amsmath}
\usepackage{booktabs}
\usepackage{multirow}

\usepackage[utf8]{inputenc}

\usepackage{microtype}

\usepackage{inconsolata}

\usepackage{graphicx}
\usepackage[most]{tcolorbox}
\tcbuselibrary{listings}

\newtcblisting[auto counter, number within=section]{promptbox}[2][]{
  enhanced,
  title={\textbf{} #2}, 
  colback=gray!5,        
  colframe=gray!60,      
  colbacktitle=gray!30,  
  coltitle=black,        
  fonttitle=\bfseries\small,
  attach boxed title to top left={xshift=3mm, yshift=-3mm},
  boxed title style={size=small, colframe=gray!60},
  sharp corners,
  top=4mm,               
  bottom=2mm,
  left=2mm, right=2mm,
  listing only,          
  listing options={      
    basicstyle=\small\ttfamily, 
    breaklines=true,            
    columns=fullflexible,       
    keepspaces=true,            
    aboveskip=0pt, belowskip=0pt
  },
  label={#1}             
}

\title{AtomMem: Building Simple and Effective Memory System \\ for LLM Agents via Atomic Facts}

\author{
  \textbf{Yanyu Yao\textsuperscript{1}},
  \textbf{Shangze Li\textsuperscript{1}},
  \textbf{Zhi Zheng\textsuperscript{1,*}},
  \textbf{Hui Zheng\textsuperscript{2}},
  \textbf{Qi Liu\textsuperscript{1}},
  \textbf{Tong Xu\textsuperscript{1,*}},
  \textbf{Enhong Chen\textsuperscript{1}}
  \\
\textsuperscript{1}State Key Laboratory of Cognitive Intelligence, \\University of Science and Technology of China, Hefei, China
\\
\textsuperscript{2}Anhui University, Hefei, China
\\
\small{
\href{mailto:yyyao@mail.ustc.edu.cn,lishangze@mail.ustc.edu.cn, liuqilq@mail.ustc.edu.cn}{\{yyyao, lishangze, liuqilq\}@mail.ustc.edu.cn}, \href{mailto:zhengzhi97@ustc.edu.cn,tongxu@ustc.edu.cn, cheneh@ustc.edu.cn}{\{zhengzhi97, tongxu, cheneh\}@ustc.edu.cn},
    \href{mailto:huizheng@ahu.edu.cn}{huizheng@ahu.edu.cn}
  }
}

\begin{document}
\raggedbottom
\maketitle

\begingroup
\renewcommand{\thefootnote}{*}
\footnotetext{Corresponding authors.}
\endgroup

\begin{abstract}
Large language models (LLMs) demonstrate strong reasoning and generation abilities, but their fixed context windows limit long-term information accumulation and reuse across multi-session interactions. Existing memory-augmented systems often construct memory in a coarse and unstable manner, relying on inefficient memory representations or unstable unconstrained updates. To address these challenges, we propose \textbf{AtomMem}, a long-term memory system designed for value-dense storage and stable memory evolution. AtomMem introduces an Atomic Fact Extractor, which selectively extracts high-value atomic facts from long-term interactions to serve as highly efficient memory representations. Subsequently, AtomMem organizes these facts into hierarchical event structures and temporal profiles, capturing coherent episodic contexts and tracking dynamically evolving user attributes over time. During retrieval, the system activates an associative memory graph to connect fragmented memories. Experiments on the LoCoMo benchmark confirm that AtomMem achieves state-of-the-art performance across various reasoning tasks, offering a scalable and economically viable solution for deploying intelligent personalized agents. The implementation code is publicly available at \url{https://github.com/MINE-USTC/AtomMem}.
\end{abstract}

\section{Introduction}

Large language models (LLMs) have demonstrated remarkable capabilities in language understanding, reasoning, and generation \cite{openai2023gpt4,bubeck2023sparks,touvron2023llama}. Recent advances have extended these models into interactive agents capable of engaging in multi-turn conversations spanning days or even months, requiring these agents to accumulate and organize useful memories. However, as these LLM-based systems are deployed in increasingly complex and long-horizon tasks, they face significant challenges related to reliability degradation \cite{liu2024lost, xiao2024streamingllm}. Constrained by fixed-length context windows, existing models often struggle to maintain coherence and accurate retrieval over extended contexts. This often results in practical failures such as forgetting user preferences, repeating previously resolved questions, or contradicting established facts.

To address this limitation, an increasing body of work has explored augmenting LLMs with external memory modules. These memory-augmented agents aim to improve long-term performance by optimizing memory management and utilization, primarily through the design of effective mechanisms for memory storage, update, and retrieval. Advanced systems like Mem0 \cite{mem0} incorporate graph databases to enhance relational organization, and A-Mem \cite{xu2025amem} enables dynamic memory evolution without predefined rules.
\begin{figure*}[t]
  \centering
  \includegraphics[width=\textwidth]{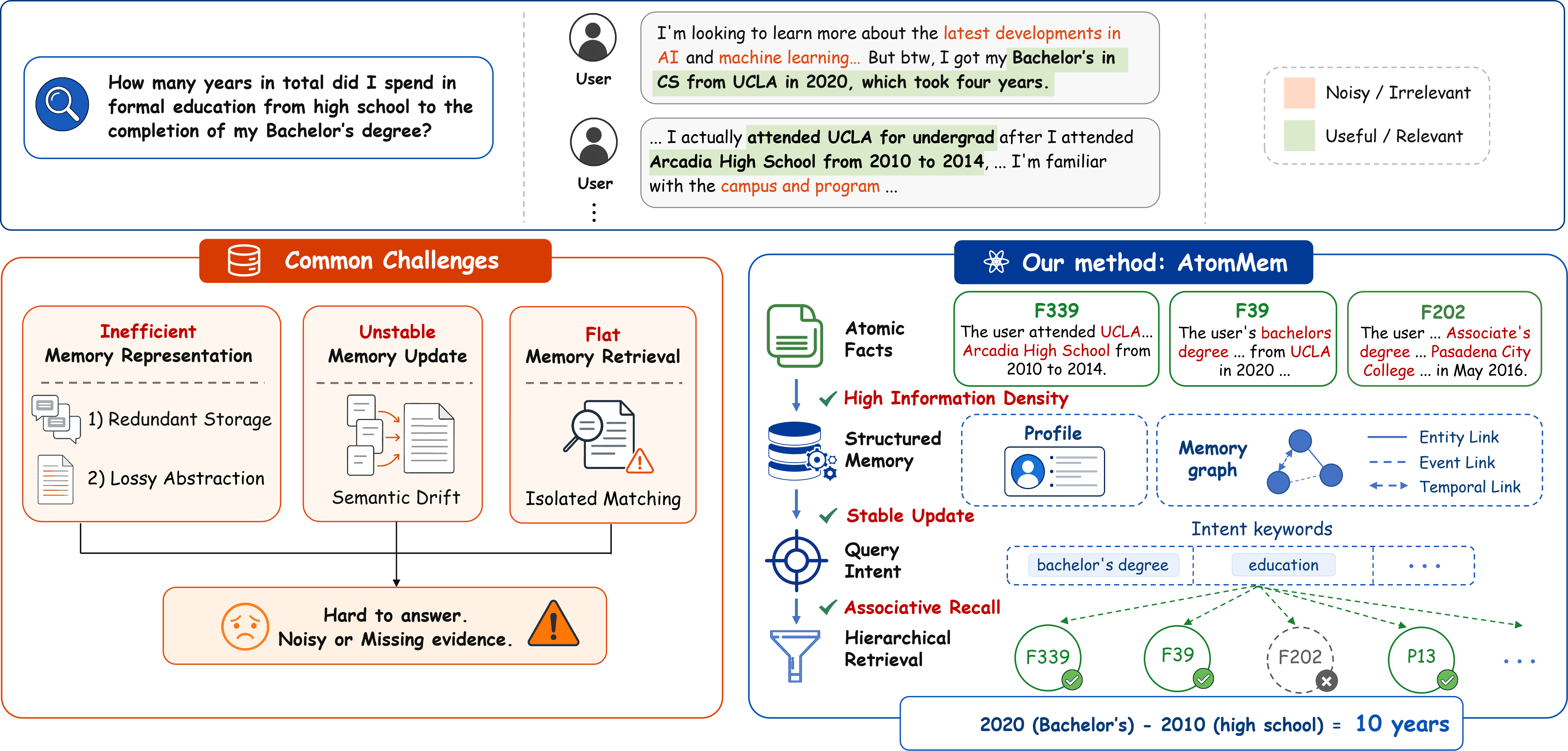}
  \caption{
    \textbf{Motivation and overview of AtomMem.} Existing memory systems face different trade-offs in memory representation, update, and retrieval. AtomMem addresses these challenges through atomic facts, constrained memory evolution, and association-graph retrieval.
    }
  \label{fig:intro}
\end{figure*}

Despite these advances, existing memory-augmented systems still face challenges in constructing reliable long-term memory due to an inherent dilemma. Storing raw conversations maximizes information retention but overwhelms retrieval-augmented generation paradigms \cite{lewis2020retrieval} with redundant noise. This bloat forces models to process irrelevant contexts. Conversely, condensed representations achieve compact formats but inevitably discard fine-grained details and accumulate noise generated by LLMs over time. Therefore, achieving a balance between high information density and contextual fidelity is crucial. A precise and reliable underlying memory representation is the fundamental prerequisite for any effective memory system.

Beyond basic representation, user memory is inherently dynamic. Preferences, experiences, and goals naturally evolve over time, requiring systems to effectively accumulate and maintain a consistent user state. Recent work has explored dynamic memory evolution, but these methods typically rely on frequent LLM-driven rewrites to update existing entries. While this design enables flexible knowledge organization and continuous adaptation, unconstrained updates introduce severe instability. Hallucinations or erroneous edits can repeatedly modify the same memory entry, leading to uncontrolled expansion and the destruction of original facts. Therefore, designing stable and controllable memory update mechanisms remains another key challenge for long-term memory systems. Furthermore, useful memories are often distributed across multiple sessions. Current memory systems often rely on flat retrieval over isolated items. This flat approach struggles to capture complex associations across sessions, failing to recover associative evidence required for personalized assistance.

In this paper, we propose AtomMem, a long-term memory system centered on atomic facts that organizes user interactions into a hierarchical memory structure and activates relevant memories through graph-based associative recall. At its core, an SFT-based Atomic Fact Extractor extracts self-contained atomic facts from raw conversations by selecting high-value information and performing lightweight reasoning such as coreference resolution and temporal anchoring. Serving as the basic semantic units of memory, these atomic facts allow AtomMem to construct event memory by associating new information with existing events or creating new ones via semantic and temporal reasoning, thereby transforming isolated facts into episodic memory. To maintain long-term user states, AtomMem builds temporal profile memory from accumulated factual evidence to incrementally track stable attributes and adapt to preference shifts while preserving historical information. During retrieval, AtomMem activates a memory graph that connects facts through entity overlap, shared events, and temporal continuity, facilitating the recall of associated memories. Together, these coordinated components enable rich yet stable memory representations, allowing LLM agents to maintain a consistent and reliable understanding of users over long-term interactions.

Our main contributions are summarized as follows:

\begin{itemize}
\item We propose AtomMem, a long-term memory framework centered on atomic facts that generates memory-aware responses through graph-based associative recall. This framework provides a stable and scalable long-term storage solution for LLM agents.
\item We introduce an atomic fact extraction module that converts noisy raw dialogues into self-contained storage units with structured metadata. This module provides a compact and faithful base representation for long-term memory. Additionally, we release a high-quality dataset to facilitate the fine-tuning of robust conversational fact extraction.
\item Comprehensive evaluations on long-term benchmarks demonstrate that AtomMem consistently outperforms state-of-the-art baselines. Notably, our simplified fact-level variant achieves competitive performance at minimal computational cost, while the full modular design yields further significant gains.
\end{itemize}

\section{Related Work}
\subsection{Retrieval-Augmented Generation}
Retrieval-Augmented Generation (RAG) combines a language model's parametric memory with an external non-parametric memory, grounding generation in retrieved evidence \cite{lewis2020retrieval}. Early work such as REALM \cite{guu2020realm} demonstrated the effectiveness of retrieval-augmented language modeling for open-domain question answering. Subsequent RAG methods improved retrieval quality through query transformation and reranking \cite{gao2023retrieval}, while FLARE and Self-RAG introduced adaptive retrieval and self-reflection \cite{jiang2023active,asai2024selfrag}. Recent studies have further explored the granularity and organization of retrieved knowledge. Dense X \cite{densex} decomposes static documents into fine-grained propositions to cover full semantics. GraphRAG and HippoRAG \cite{graphrag,hipporag} organize document corpora into entity- or concept-centered knowledge graphs for retrieval. In LLM agents, retrieval-based access has become an important mechanism for exposing external knowledge and long-term memory to the agent \cite{park2023generative,xi2023rise}.

\subsection{Memory for LLM Agents}
The architectural design of memory-augmented LLM agents primarily depends on their memory abstraction. Some systems represent memory as textual experiences or higher-level reflections. For example, Think-in-Memory \cite{tim2023} stores evolving historical thoughts, and RMM \cite{tan2025prospect} dynamically summarizes dialogue history across granularities. A second category focuses on symbolic or relational memory. This approach anchors information on structured objects, such as triplets in RET-LLM \cite{modarressi2023retllm}, and knowledge graphs in Mem0 \cite{mem0}. Furthermore, frameworks like MemGPT \cite{packer2023memgpt} and MemoryOS \cite{kang2025memoryosaiagent} manage memory through explicit hierarchical interfaces. Recent studies such as A-Mem \cite{xu2025amem}, MEM1 \cite{zhou2025mem1}, and MemCoE \cite{xu2026memcoe} further explore self-organizing or learned memory strategies. For comprehensive evaluation, benchmarks such as LoCoMo \cite{maharana2024evaluating} and LongMemEval \cite{wu2024longmemeval} assess long-term conversational memory, while PERMA~\cite{liu2026perma} and VehicleMemBench~\cite{chen2026vehiclemembench} provide complementary benchmarks for evaluating evolving user preferences and multi-user long-term memory, respectively. While existing methods often rely on coarse or lossy memory representations, retrieved memories may remain ambiguous and difficult to interpret or use in isolation. In contrast, AtomMem uses self-contained atomic facts as a compact and faithful memory representation, and organizes them through a dynamic association graph to connect related evidence and support complex retrieval across long-term interactions.

\section{Methods}
\begin{figure*}[t]
  \centering
  \includegraphics[width=0.95\textwidth]{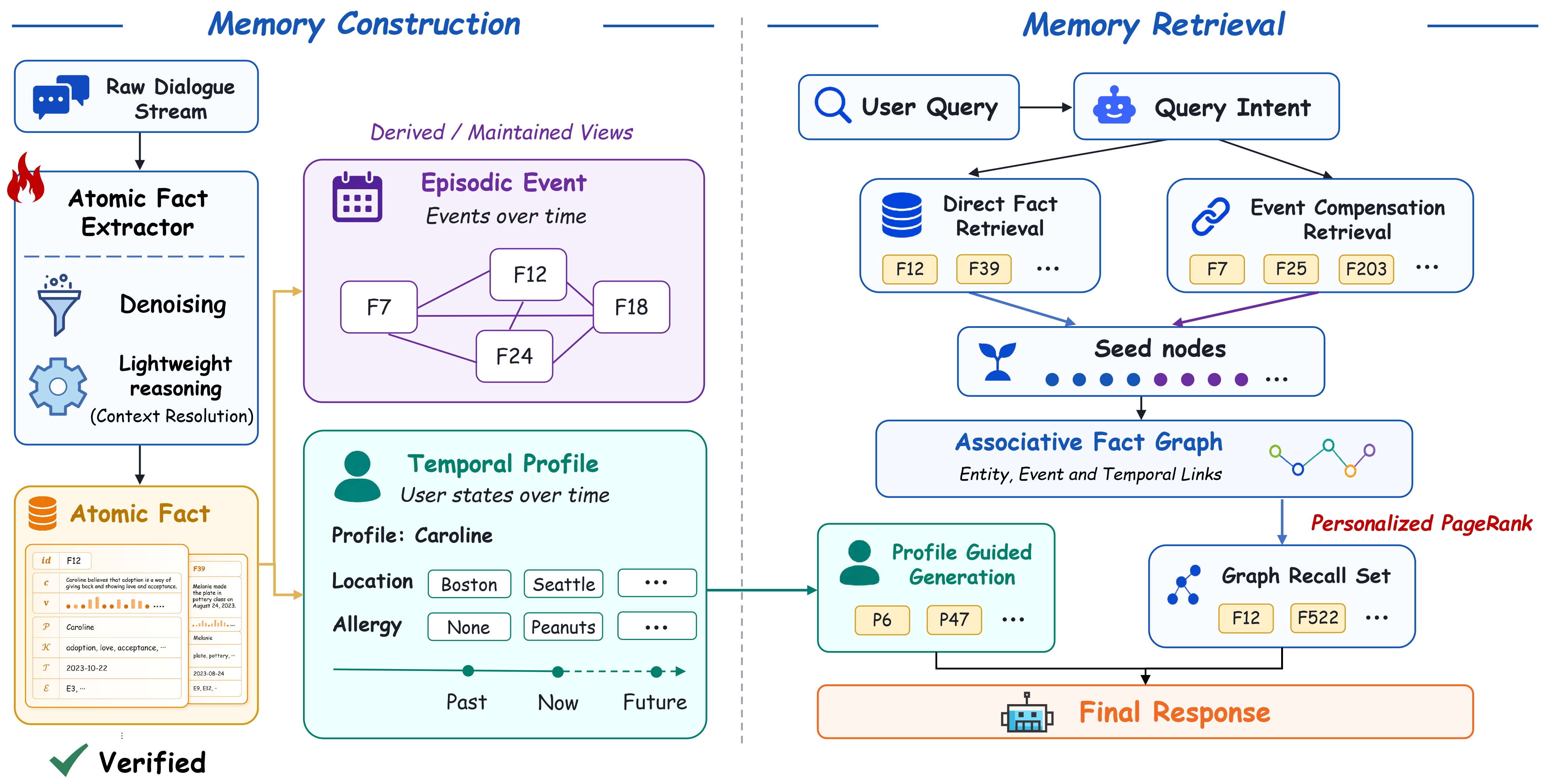}
  \caption{
    \textbf{The overall architecture of AtomMem.} It is designed to support high-information-density memory storage, stable user-state evolution, and efficient retrieval for long-term personalized agents.
    }
  \label{fig:k_sensitivity}
\end{figure*}
AtomMem is designed to transform unstructured dialogue streams into a structured and retrieval-friendly memory representation. It first extracts self-contained atomic facts and consolidates related facts into contextual event memory while dynamically modeling user states with temporal profiles. During retrieval, AtomMem activates related memories through a memory graph and integrates the activated memories to generate the final response.

\subsection{Base Representation: Atomic Fact Extraction}
As interactions lengthen, useful information is sparsely distributed across noisy dialogue turns. Moreover, these raw dialogues frequently rely on implicit context such as unresolved pronouns (e.g., "he", "it") and relative temporal references (e.g., "last Friday"), which become ambiguous when retrieved in isolation. We define an atomic fact as a minimal, self-contained unit of high-value information that expresses a single semantically coherent proposition. It remains faithful to the original dialogue and resolves context-dependent references, allowing it to be understood and used independently of the source context. Accordingly, our goal is to transform raw dialogue sessions into a set of structured atomic facts $F$, thereby providing a superior memory representation.
\subsubsection{Atomic Fact Extractor}
\label{fact-extract}
To address redundancy and noise, we introduce an Atomic Fact Extractor trained via supervised fine-tuning (SFT) that performs essential denoising and lightweight reasoning such as coreference resolution. Since generating high-quality atomic facts requires complex reasoning, relying solely on heuristic rules or zero-shot prompting often yields suboptimal results. To overcome this, we construct a high-quality dataset $\mathcal{D}$ (see Appendix~\ref{stf_detail} for data construction details) through a two-stage data construction pipeline. 

We train a lightweight LLM using the constructed dataset. Formally, given the instruction $I$ and the dialogue context $C$, we optimize the model parameters $\theta$ to maximize the likelihood of the target atomic facts $F$:
\begin{equation}
  \label{eq:fact_extractor_objective}
  \max_{\theta}
  \sum_{(I,C,F)\in\mathcal{D}}
  \log P_{\theta}(F \mid I, C)
\end{equation}
By fine-tuning on this specialized distribution, the model acts as an efficient information filter. It compresses raw interactions into a dense representation before they enter the memory system, while ensuring that each generated fact is independent and comprehensible without external context.

\subsubsection{Structured Fact Construction}
While the Atomic Fact Extractor provides clean textual content, efficient retrieval and memory management require structured metadata. We therefore encapsulate the extracted text into a Structured Atomic Fact, serving as the minimum semantic unit of the memory system. Formally, we define an atomic fact $F$ as:
$$F = \{id, c, \mathbf{v}, \mathcal{P}, \mathcal{K}, \mathcal{T}, \mathcal{E}\}$$
where $id$ is the fact identifier, $c$ is the self-contained text generated by the extractor, and $\mathbf{v}$ denotes its dense semantic embedding. To obtain precise symbolic metadata, we leverage the LLM to parse the conversation and extract the following key attributes. Specifically, $\mathcal{P}$ denotes the participants involved in the interaction, $\mathcal{K}$ represents the topical keywords, and $\mathcal{T}$ captures the temporal information that anchors each fact to a specific timestamp or interval. Additionally, $\mathcal{E}$ is a list of associated event IDs, which is initialized as empty and links the fact to higher-level event blocks.

\subsubsection{Fact Verification}
Before fact storage, AtomMem verifies whether the newly generated fact $F_{new}$ duplicates or conflicts with existing records. To narrow the search space and ensure contextual relevance, we first construct a candidate set by filtering the global memory based on shared metadata such as participants and temporal contexts. This symbolic filtering effectively eliminates irrelevant facts before expensive vector computations. To retrieve the top-$k$ items from this candidate set, we define a universal hybrid similarity metric combining semantic embedding and keyword Jaccard similarities for any two inputs $x$ and $y$:
\begin{equation}
  \label{eq:hybrid_score}
  S_{h}(x,y)
  = \alpha\ \cdot \operatorname{sim}_{e}(\mathbf{v}_x,\mathbf{v}_y)
  + \beta\ \cdot\operatorname{Jac}(\mathcal{K}_x,\mathcal{K}_y)
\end{equation}
where $\mathbf{v}$ and $\mathcal{K}$ denote vector embeddings and keyword sets while $\alpha$ and $\beta$ balance semantic similarity and keyword overlap. AtomMem then ranks the candidates using $S_{h}(F_{new}, F_i)$ in Eq.~\ref{eq:hybrid_score} and selects the top $k$ facts $\mathcal{C}_{ret} = \{F_1, ..., F_k\}$ with the highest scores as the relevant context.

Based on the retrieved candidates, the LLM analyzes the relationship between the new input $c_{new}$ and the retrieved context $\mathcal{C}_{ret}$ to generate precise content for storage or update. We formalize the verification as a function mapping the new input and context to a residual content and a set of updates:
$$(c'_{new}, \mathcal{U}) \leftarrow \text{LLM}(c_{new} \parallel \mathcal{C}_{ret})$$
where $c'_{new}$ represents the residual novel information not entailed by the existing context. The system stores this non-redundant content as a new atomic fact. Additionally, $\mathcal{U}$ denotes a set of update tuples for existing facts, generated only when logical conflicts are detected. This joint mechanism effectively prevents memory redundancy and dynamically maintains global consistency.

\subsection{Episodic Consolidation: Event Memory Construction}
While atomic facts provide precise details, they lack the contextual continuity of broader experiences. Therefore, we structure memory into Events, which aggregate related facts into coherent narrative blocks. Formally, an Event $E$ is defined as:

$$E = \{id, \mathcal{S}, \mathcal{F}_{ids}, \mathcal{P}_{e}, \mathcal{K}_{e}, \mathcal{T}_{e}\}$$
where $\mathcal{S}$ is the concise summary, $\mathcal{F}_{ids}$ is the set of constituent fact IDs, and $\mathcal{P}_{e}$, $\mathcal{K}_{e}$, and $\mathcal{T}_{e}$ describe the participants, keywords, and temporal span of the event. To maintain this structure, AtomMem dynamically absorbs new logically aligned facts into existing events or triggers the creation of new ones. The complete event construction and update algorithm is detailed in Appendix~\ref{event-memory}.



\subsection{State Evolution: Temporal Profile Modeling}
Beyond episodic events, understanding users requires modeling stable long-term attributes such as preferences, habits, and background information. Therefore, we introduce a profile layer to capture these persistent yet dynamic user states. Formally, a profile entry $P$ is structured as follows: 
$$P = \{id, u, c, v_p, \mathcal{K}_p, \mathcal{E}_{evi}, t_{from}, \mathcal{H}\}$$
where $u$ denotes the user, $c$ describes the stable attribute of the user, $v_p$ represents the vector embedding of the content, $\mathcal{K}_p$ is the keyword set, $t_{from}$ is the effective timestamp, $\mathcal{H}$ stores historical versions, and $\mathcal{E}_{evi}$ tracks supporting fact IDs for traceability.

Unlike the real-time updates for facts and events, profile construction employs a session-based batch mechanism. During atomic fact extraction, the LLM identifies facts implying potential long-term characteristics and temporarily adds them to a waiting queue. At the end of a dialogue session, the system processes these queued candidate facts in a batch to generate structured profiles. 

For each queued candidate, AtomMem retrieves the top-$k_p$ most relevant existing profiles for the same user $u$ using $S_{h}(P_{new}, P_i)$ in Eq.~\ref{eq:hybrid_score}. An LLM-driven updater then decides whether to mark the candidate as redundant, update the current profile, modify a historical version, or create an entirely new entry. When the current profile changes, AtomMem copies the previous state into the history $\mathcal{H}$ and records its valid time interval or updates the corresponding historical version. This mechanism allows the system to accumulate stable preferences while preserving past user states.
\subsection{Associative Recall: Memory Graph Activation}
\label{fact_graph}
Long-term memory often relies on associative recall. These useful memories naturally connect through past events, shared topics, or adjacent dialogue contexts. Therefore, AtomMem activates a memory graph over the atomic facts. This graph uses atomic facts as nodes and encodes three distinct types of associations as edges.

\paragraph{Entity Edge} Two facts connect if they share keywords. To mitigate the noisy connections introduced by frequent keywords, AtomMem calculates a local edge weight between two facts $F_i$ and $F_j$ using an IDF-weighted overlap:
\begin{equation}
  \label{eq:keyword_edge_weight}
  w_{\mathrm{ent}}(F_i, F_j)
  =
  \frac{
    \sum_{k \in \mathcal{K}_i \cap \mathcal{K}_j}
    \omega(k)
  }{
    \sqrt{
      \sum_{k \in \mathcal{K}_i}
      \omega(k)
      \sum_{k \in \mathcal{K}_j}
      \omega(k)
      +
      \epsilon
    }
  }
\end{equation}
where $\mathcal{K}_i$ and $\mathcal{K}_j$ are the respective keyword sets. The query-aware weight $\omega(k)$ boosts query-relevant keywords and penalizes frequent non-informative terms, while $\epsilon$ is a small constant added to prevent division by zero and ensure numerical stability.

\paragraph{Event Edge} Two facts connect when they belong to the same event. This edge allows the system to recall facts based on a coherent episodic background even if they lack keyword similarity. To reduce noise from overly broad events, the edge weight incorporates a penalty based on event size:
\begin{equation}
  \label{eq:event_edge_weight}
  w_{\mathrm{event}}(F_i, F_j)
  =
  \sum_{e \in \mathcal{E}_i \cap \mathcal{E}_j}
  \frac{1}
  {\left(|\mathcal{F}_e| - 1\right)^{\gamma_e}}
\end{equation}
where $\mathcal{E}_i$ and $\mathcal{E}_j$ denote the event sets for the two facts, and $\mathcal{F}_e$ is the set of member facts within event $e$. The penalty coefficient $\gamma_e$ controls the edge weight decay rate for large events.

\paragraph{Temporal Edge} Two facts connect when they appear in adjacent dialogue turns within the same session. The edge weight decays according to the turn distance:
\begin{equation}
  \label{eq:turn_edge_weight}
  w_{\mathrm{temp}}(F_i, F_j)
  =
  \exp
  \left(
  -
  \frac{
    |\mathrm{pos}_i - \mathrm{pos}_j|
  }{
    \tau
  }
  \right)
\end{equation}
where $\mathrm{pos}_i$ and $\mathrm{pos}_j$ denote dialogue positions, and the decay coefficient $\tau$ controls the impact of turn distance. AtomMem restricts temporal edges to intra-dialogue facts within a maximum window $W$ ($|\mathrm{pos}_i - \mathrm{pos}_j| \le W$).
\subsection{Response Generation: Hierarchical Memory Integration}
To ensure that the agent responds with accurate and contextually complete information, we design a hierarchical retrieval mechanism. This process transforms a user's natural language query into a structured search command, executes a multi-strategy recall, and synthesizes the final response.
\subsubsection{Query Intent Analysis}
The retrieval process begins by analyzing the user's input query $q$ to determine the specific information needs. We employ an LLM to parse $q$ and extract a structured query object $Q$, defined as: 
$$Q = \{\mathrm{I}_{prof}, \mathcal{P}_q, \mathcal{K}_q, \mathcal{T}_q\}$$where $\mathrm{I}_{prof} \in \{0, 1\}$  indicates whether user profiles are required, $\mathcal{P}_q$ denotes the involved participants, $\mathcal{K}_q$ captures the core intent via extracted keywords, and $\mathcal{T}_q$ specifies the relevant time range if any.
\subsubsection{Hierarchical Hybrid Retrieval}

To balance precision and contextual breadth, we implement a hierarchical retrieval strategy consisting of three stages.

(1) \textbf{Primary Recall}: AtomMem first filters global facts using participant ($\mathcal{P}_q$) and temporal ($\mathcal{T}_q$) constraints. We evaluate the remaining candidates against the query $Q$ using the predefined metric $S_{h}(F, Q)$ in Eq.~\ref{eq:hybrid_score}. The top $\frac{k_s}{2}$ facts form the primary set $\mathcal{R}_{pri}$.

(2) \textbf{Compensatory Recall}: Direct retrieval often misses implicit yet relevant context. Therefore, AtomMem evaluates the event layer. After applying metadata filters, the system ranks events using $S_{h}(E, Q)$ in Eq.~\ref{eq:hybrid_score} to identify the top events. We extract all constituent facts from these events into a candidate pool. We exclude items already in $\mathcal{R}_{pri}$ to avoid redundancy. AtomMem then ranks these candidates using a fusion score:
$$S_f(F) = w_e \cdot S_{h}(E, Q) + w_f \cdot S_{h}(F, Q)$$
where $w_e$ and $w_f = 1-w_e$ denote the event-level and fact-level weights, respectively, balancing global event relevance and local fact precision. The top $\frac{k_s}{2}$ facts from this pool form the compensatory set $\mathcal{R}_{comp}$. The combined set $\mathcal{R}_{seed} = \mathcal{R}_{pri} \cup \mathcal{R}_{comp}$ ensures both direct precision and contextual completeness.

(3) \textbf{Associative Recall}: We use $\mathcal{R}_{seed}$ as seeds to activate the memory graph. AtomMem constructs a localized graph around these seeds by expanding limited hops and retaining only top neighbors  that satisfy participant and temporal constraints. AtomMem then applies Random Walk with Restart (RWR) to propagate activations across entity, event and temporal edges. Finally, the system selects the top $k_f$ facts with the highest activation scores across all nodes to output the final retrieved context $\mathcal{R}_{fact}$.

\subsubsection{Profile Augmentation and Response Generation}
If the intent analysis sets the flag $\mathrm{I}_{prof}=1$, AtomMem executes profile recall to retrieve stable user attributes. The system first filters the global repository to retain profiles matching the query participants $\mathcal{P}_q$ and then ranks these candidates using the hybrid similarity metric $S_{h}(\mathcal{P}_i, Q)$ in Eq.~\ref{eq:hybrid_score}. The top-$k_p$ items form the profile context set $\mathcal{R}_{prof}$. If the input contains temporal constraints, the system selects profile versions valid during that specific time. This ensures the response reflects both current user preferences and historical user states.

Finally, AtomMem constructs a comprehensive context $\mathcal{C}$ by concatenating the retrieved episodic fact set $\mathcal{R}_{fact}$ and the semantic profile set $\mathcal{R}_{prof}$. This context $\mathcal{C}$, along with the original query $q$, is fed into the LLM to generate the final response. This multi-source injection mechanism ensures that the agent's answer is grounded in specific atomic details, enriched by event-level context, and personalized by long-term user attributes.
\section{Experiments}
\label{sec:experiments}
\begin{table*}[t]
  \centering
  \small
  \setlength{\tabcolsep}{4.5pt}
  \renewcommand{\arraystretch}{1.08}
  \resizebox{\textwidth}{!}{
  \begin{tabular}{lccccccccccc}
  \toprule
  \multirow{2}{*}{\textbf{\textsc{Method}}}
  & \multicolumn{3}{c}{\textsc{Single-Hop}}
  & \multicolumn{3}{c}{\textsc{Multi-Hop}}
  & \multicolumn{3}{c}{\textsc{Temporal}}
  & \textsc{Open-Domain}
  & \multirow{2}{*}{\textsc{Tokens(K)}} \\
  \cmidrule(lr){2-4}
  \cmidrule(lr){5-7}
  \cmidrule(lr){8-10}
  \cmidrule(lr){11-11}
  & $F_1 \uparrow$ & BLEU-1 $\uparrow$ & $J \uparrow$
  & $F_1 \uparrow$ & BLEU-1 $\uparrow$ & $J \uparrow$
  & $F_1 \uparrow$ & BLEU-1 $\uparrow$ & $J \uparrow$
  & $J \uparrow$
  & \\
  \midrule
    \textsc{LoCoMo}
    & 37.81 & 26.23 & 57.07
    & 20.97 & 11.09 & 48.58
    & 34.79 & 15.46 & 38.01
    & 41.67
    & 827.20 \\

    \textsc{MemoryBank}
    & 17.85 & 12.13 & 46.61
    & 11.25 & 8.20 & 28.01
    & 14.26 & 9.60 & 34.27
    & 23.96
    & 926.98 \\

    \textsc{A-Mem}
    & 33.87 & 27.64 & 54.10
    & 21.17 & 15.40 & 43.26
    & 34.13 & 29.98 & 34.89
    & 32.29
    & 11687.58 \\

    \textsc{MEM0}
    & \underline{54.95} & \underline{44.71} & \underline{78.00}
    & 36.02 & 25.25 & 62.41
    & 30.36 & 29.69 & 30.53
    & \underline{54.17}
    & 55300.30 \\

    \textsc{MemoryOS}
    & 49.72 & 43.58 & 66.47
    & \underline{37.15} & 26.87 & 60.99
    & 41.99 & 36.32 & 34.58
    & 51.04
    & 19207.67 \\

    \textsc{LightMem}
    & 49.30 & 42.45 & 68.97
    & 36.59 & 27.27 & \underline{64.89}
    & 47.41 & 41.29 & 51.09
    & 43.75
    & 5021.56 \\

    \midrule
    \textsc{AtomMem-Flat}
    & 47.08 & 40.40 & 67.66
    & 37.03 & \underline{29.50} & 55.67
    & \underline{56.45} & \underline{48.98} & \underline{59.50}
    & 52.08
    & 722.75 \\

    \textsc{AtomMem}
    & \textbf{56.66} & \textbf{49.56} & \textbf{78.48}
    & \textbf{42.50} & \textbf{33.26} & \textbf{68.44}
    & \textbf{62.78} & \textbf{57.64} & \textbf{66.98}
    & \textbf{64.58}
    & 21357.06 \\
    \bottomrule
  \end{tabular}
  }
  \caption{
    Performance comparison of various memory methods on the LoCoMo benchmark. The best and second-best results are bolded and underlined.
  }
  \label{tab:memory_comparison}
\end{table*}

\subsection{Experimental Settings}
\paragraph{Datasets} We evaluate AtomMem on the LoCoMo \cite{maharana2024evaluating} and LongMemEval \cite{wu2024longmemeval} benchmarks. LoCoMo is a challenging dataset widely adopted for assessing long-context capabilities in memory-augmented LLMs. Designed to test the management of remote dependencies and consistency, LoCoMo consists of long-term interactions that average over 600 turns across 35 sessions, paired with comprehensive question sets. Additionally, LongMemEval specifically targets the interactive memory capabilities of chat assistants by using 500 curated questions to evaluate diverse functions. The experimental results on LongMemEval are detailed in Appendix~\ref{longmemeval}.

\paragraph{Baselines and Metrics} To validate the effectiveness of our approach, we benchmark AtomMem against representative memory modeling systems, including LoCoMo \cite{maharana2024evaluating}, MemoryBank \cite{zhong2023memorybank}, A-MEM \cite{xu2025amem}, MEM0 \cite{mem0}, MemoryOS \cite{kang2025memoryosaiagent} and LightMem \cite{fang2026lightmem}. For fair comparison, we re-implemented all baselines using GPT-4o-mini as the uniform backbone model. Performance is measured using three complementary metrics: Token-level F1 Score (F1), BLEU-1, and LLM-as-a-Judge (J). The LLM-as-a-Judge metric employs DeepSeek-V4-Pro with a rigorous evaluation prompt to assess semantic correctness, providing a closer approximation to human judgment. Additionally, we track the total API token consumption across the entire pipeline to evaluate cost efficiency. 

\paragraph{Experimental Details} We fine-tuned a Qwen3-14B model as our fact extractor using SFT with LoRA on a single NVIDIA A100 GPU. The training configuration included a learning rate of 5e-5, a LoRA rank of 128, and an effective global batch size of 8 over 3 epochs. To ensure consistent experimental conditions, we employed all-MiniLM-L6-v2 as the unified text embedding model for both our approach and all baseline systems. Furthermore, we standardized the retrieval capacity by setting the top-$k$ to 10 for all comparative baselines and our method. Detailed hyperparameter configurations for AtomMem's hierarchical retrieval and fusion mechanisms are provided in Appendix~\ref{app:hyperparameters}.

\subsection{Main Results}

\begin{table*}[t]
  \centering
  \small
  \setlength{\tabcolsep}{3.8pt}
  \renewcommand{\arraystretch}{1.08}
  \newcommand{\best}[1]{\textbf{#1}}
  \resizebox{\textwidth}{!}{
  \begin{tabular}{l*{3}{cccc}cc}
    \toprule
    \multirow{2}{*}{\textbf{\textsc{Method}}}
    & \multicolumn{4}{c}{\textsc{Single-Hop}}
    & \multicolumn{4}{c}{\textsc{Multi-Hop}}
    & \multicolumn{4}{c}{\textsc{Temporal}}
    & \multicolumn{2}{c}{\textsc{Open-Domain}} \\
    \cmidrule(lr){2-5}
    \cmidrule(lr){6-9}
    \cmidrule(lr){10-13}
    \cmidrule(lr){14-15}
    & $F_1$ & BLEU-1 & R@10 & $J$
    & $F_1$ & BLEU-1 & R@10 & $J$
    & $F_1$ & BLEU-1 & R@10 & $J$
    & R@10 & $J$ \\
    \midrule
    \textsc{LoCoMo}
    & 37.81 & 26.23 & 56.98 & 57.07
    & 20.97 & 11.09 & 32.91 & 48.58
    & 34.79 & 15.46 & 56.10 & 38.01
    & 33.55 & 41.67 \\

    \midrule
    \textsc{AtomMem-FLAT}
    & 47.08 & 40.40 & 72.18 & 67.66
    & 37.03 & 29.50 & 41.75 & 55.67
    & 56.45 & 48.98 & 79.89 & 59.50
    & 42.46 & 52.08 \\

    w/o Fact Verification
    & 55.17 & 48.48 & 75.86 & 77.05
    & 40.67 & 32.01 & 46.08 & 65.96
    & 62.04 & 56.94 & 80.37 & 65.73
    & 47.11 & 61.46 \\

    w/o Event
    & 53.25 & 46.46 & 75.56 & 72.29
    & 37.12 & 29.77 & 42.33 & 56.38
    & 58.81 & 54.05 & 77.28 & 64.17
    & 41.81 & 51.04 \\

    w/o Profile
    & 50.91 & 44.59 & 75.21 & 68.73
    & 38.33 & 30.29 & 44.56 & 59.22
    & 57.77 & 51.44 & 79.00 & 62.93
    & 44.96 & 54.17 \\

    w/o Entity Edge
    & 52.76 & 46.31 & 75.14 & 73.37
    & 40.11 & 30.77 & 46.95 & 64.18
    & 57.85 & 52.84 & 78.37 & 63.23
    & 44.10 & 60.42 \\

    w/o Event Edge
    & 53.34 & 46.82 & 75.76 & 72.77
    & 40.73 & 31.63 & 47.23 & 64.89
    & 59.06 & 54.29 & 78.27 & 64.17
    & 44.88 & 57.29 \\

    w/o Temporal Edge
    & 53.35 & 46.81 & 75.54 & 73.72
    & 40.84 & 31.49 & 47.29 & 64.89
    & 58.50 & 53.23 & 79.02 & 64.49
    & 45.23 & 59.38 \\

    w/o Graph
    & 50.55 & 44.14 & 74.08 & 71.82
    & 39.76 & 30.91 & 46.72 & 62.76
    & 60.90 & 54.74 & 80.33 & 62.93
    & 47.73 & 60.42 \\

    \midrule
    \textsc{AtomMem}
    & \best{56.66} & \best{49.56} & \best{76.30} & \best{78.48}
    & \best{42.50} & \best{33.26} & \best{48.15} & \best{68.44}
    & \best{62.78} & \best{57.64} & \best{81.10} & \best{66.98}
    & \best{48.98} & \best{64.58} \\
    \bottomrule
  \end{tabular}
  }
  \caption{
    Ablation study of AtomMem on LoCoMo using GPT-4o-mini.
    R@10 denotes Recall@10. The best results are highlighted in bold.
    ``AtomMem-FLAT'' removes the hierarchical memory structure.
    ``w/o Fact Verification'' removes fact verification.
    ``w/o Event'' removes episodic events.
    ``w/o Profile'' removes profile memory.
    ``w/o Entity/Event/Temporal Edge'' removes specific edges.
    ``w/o Graph'' removes graph-based associative recall.
  }
  \label{tab:ablation_study}
\end{table*}

\paragraph{Superior Performance}As presented in Table~\ref{tab:memory_comparison}, AtomMem achieves state-of-the-art results across all evaluated metrics. While maintaining a steady lead in Single-Hop tasks, our system demonstrates its distinct advantage in scenarios requiring long-context integration and complex reasoning. Specifically, in Multi-Hop and Temporal tasks, AtomMem achieves substantial improvements over the strongest baseline, LightMem, increasing the $J$ score by 5.5\% and 31.1\%, respectively. AtomMem also dominates the Open-Domain task by raising the $J$ score to 64.58 compared to 54.17 for MEM0. This consistent enhancement confirms that our hierarchical memory architecture successfully bridges the semantic gap between queries and distant historical context. AtomMem traces implicit narrative connections beyond surface similarity to ensure accurate retrieval of subtle information.
\paragraph{Computational Efficiency} AtomMem demonstrates competitive cost efficiency alongside its robust reasoning capabilities. Despite the overhead associated with memory construction, our system significantly reduces total token consumption by approximately 61.4\% compared to the highly performant Mem0. This efficiency stems from our strategy of compressing redundant conversational streams into compact, high-value atomic units, which prevents the context window from being flooded by low-utility tokens during retrieval. AtomMem strikes an optimal balance between resource consumption and advanced reasoning to offer a scalable solution for practical deployment.
\paragraph{Validation of Atomic Fact Extraction} We validate the effectiveness of our core extraction mechanism using the AtomMem-Flat variant. This simplified version lacks hierarchical event structures and relies solely on retrieving atomic facts. Despite this simplicity, AtomMem-Flat significantly outperforms the standard LoCoMo baseline operating on raw dialogue history. For instance, on the challenging Multi-Hop reasoning tasks, it raises the F1 score from 20.97 to 37.03, yielding a 76.6\% relative improvement. Critically, AtomMem-Flat achieves this performance while using the lowest number of tokens (722.75k) among all methods, comparable to LoCoMo yet delivering performance competitive with the strongest existing baseline. This result provides compelling evidence that memory storage quality is paramount. Sophisticated system orchestration cannot compensate for flawed memory representations or information loss.

\subsection{Ablation Study}
To assess the contribution of each component, we compare the full AtomMem system with multiple ablated variants, as shown in Table~\ref{tab:ablation_study}. The AtomMem-Flat variant uses simple flat retrieval without hierarchical structures but significantly outperforms the standard LoCoMo baseline. This proves that structured atomic facts provide a superior memory representation. The full AtomMem achieves the best performance across all reported metrics, indicating that each component contributes positively to the complete system. Removing Fact Verification produces modest but consistent decreases, supporting its role in maintaining memory quality. The event and profile layers are also important for capturing coherent episodic contexts and persistent user attributes. Removing events reduces the Multi-Hop $J$ score from 68.44 to 56.38, while removing profiles lowers the Single-Hop $F_1$ score from 56.66 to 50.91. Furthermore, removing any individual edge type degrades performance, suggesting that these edge types provide complementary association signals. Removing graph-based associative recall further degrades performance on complex reasoning tasks, confirming the benefit of relational propagation over isolated memory retrieval.

\subsection{Hyperparameter Analysis}
\label{ex_set}
We evaluate the retrieval capacity $k_f$ below. Additional analyses are detailed in Appendix~\ref{sec:hyperparameter_analysis}.
\paragraph{Impact of Retrieval Capacity $k_f$} We evaluated performance by varying the number of retrieved atomic facts from 5 to 40. As shown in Figure~\ref{fig:k_sensitivity}, increasing $k_f$ initially yields significant gains for all tasks, particularly in reasoning-intensive tasks. This confirms that a broader context provides necessary evidence for complex deduction. Despite continuous increases in Recall, system performance drops when $k_f$ is greater than 20. This indicates that excessive irrelevant noise degrades the reasoning quality of the LLM. Consequently, we selected a moderate setting ($k_f=10$) for our main experiments to strike an optimal balance between retrieval accuracy, token consumption, and response latency.
\begin{figure}[t]
  \includegraphics[width=\columnwidth]{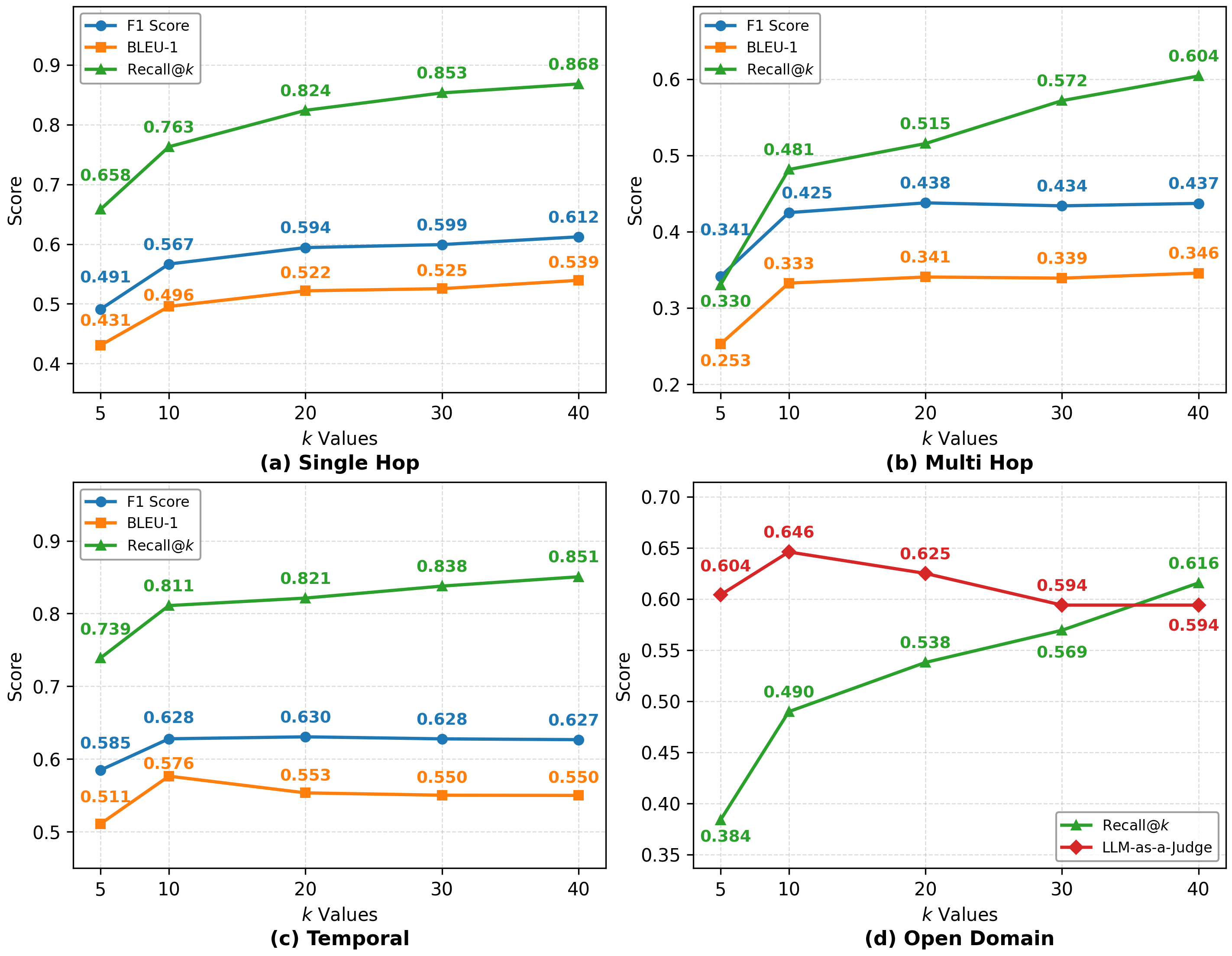}
  \caption{Performance sensitivity analysis for the memory retrieval capacity $k_f$.}
  \label{fig:k_sensitivity}
\end{figure}

\section{Conclusion}
We introduced AtomMem as a robust long-term memory framework. It distills raw dialogues into precise atomic facts to serve as a highly efficient memory representation. The system organizes these facts into hierarchical event structures and temporal profiles. This design captures coherent episodic contexts and tracks dynamically evolving user attributes. AtomMem then activates an associative memory graph to connect scattered facts during retrieval. Experiments on the LoCoMo benchmark confirm that AtomMem achieves state-of-the-art performance. The superior results of our simplified flat variant further validate that memory storage quality is paramount. This proves that structured facts provide a fundamentally better retrieval basis than unstructured history. Ultimately, AtomMem offers a scalable and economically viable solution to deploy intelligent agents capable of sustaining personalized long-term interactions.

\section*{Acknowledgments}

This work was supported in part by grants from the New Generation Artificial Intelligence-National Science and Technology Major Project (No.~2025ZD0122701), the Anhui Natural Science Foundation (No.~2508085ZD006), the National Natural Science Foundation of China (No.~62502480), the Postdoctoral Fellowship Program and the China Postdoctoral Science Foundation under Grant Nos.~BX20250387 and 2025M781529, and the Fundamental Research Funds for the Central Universities under Grant No.~WK2150250042.

\clearpage
\section*{Limitations}
\noindent\textbf{Reliance on base model performance.} Multiple stages in our framework rely on the underlying LLM, making overall performance sensitive to its generation quality and output consistency.
\par\vspace{\baselineskip}

\noindent\textbf{Lack of multimodal support.} The current system processes only textual interactions, whereas real-world conversations often involve multimodal inputs such as images and audio. Extending the system to multimodal interactions remains an important direction for future research.
\par\vspace{\baselineskip}

\noindent\textbf{Further token efficiency optimization.} Although AtomMem achieves a favorable balance between token efficiency and system performance, its overall token consumption could be further optimized.

\clearpage
\bibliography{main}

\clearpage
\appendix

\section{Appendix}
\setcounter{figure}{0}
\setcounter{table}{0}
\counterwithin{table}{section}
\counterwithin{figure}{section}
\label{sec:appendix}
\subsection{Details of Dataset Construction}
\label{stf_detail}
In this section, we provide a detailed description of the data construction process for our instruction-tuning dataset $\mathcal{D}$. This includes the source of the raw dialogues, the specific guidelines used for fact extraction, and the final dataset statistics.
\subsubsection{Data Source and Diversity}
We utilized the LoCoMo generation pipeline to produce high-quality synthetic dialogues. This pipeline uses LLM-based agents guided by character profiles and temporal event graphs, ensuring that the generated conversations remain consistent over long-term interactions.

We meticulously designed 10 fictional character groups spanning diverse ages, occupations, personalities, cultural backgrounds, and life stages. To prevent data leakage, all characters, event graphs, and dialogue instances used for SFT are distinct from those in the LoCoMo evaluation set. The resulting dataset covers diverse conversational topics and realistic scenarios, ranging from routine activities to significant life events. The complete character profiles, event graphs, and generated dialogues are publicly available in our GitHub repository.\footnote{\url{https://github.com/MINE-USTC/AtomMem}}

\subsubsection{Extraction Pipeline and Prompt Guidelines} 
We employed a two-stage pipeline to construct the dataset: Teacher-Model Preliminary Extraction followed by Human-Guided Refinement.

(1) \textbf{Stage 1: Teacher-Model Extraction}

We utilized GPT-4o as the teacher model to perform the initial extraction. The model was provided with specific guidelines to filter noise and resolve contextual dependencies. The exact system prompt used is presented in Figure \ref{prompt:teacher_rules}.

(2) \textbf{Stage 2: Human-Guided Refinement}

Human annotators reviewed all samples generated by the teacher model to guarantee high data quality. Rigorous manual verification confirmed comprehensive denoising, unambiguous coreference resolution, the decomposition of long sentences, and the elimination of hallucinations. Their primary responsibilities were:
\begin{itemize}
    \item Verification: Checking whether the extracted facts accurately reflected the original dialogue without hallucination.
    \item Context Resolution: Correcting any unresolved pronouns or vague temporal references that the teacher model missed.
    \item Filtering: Removing any residual conversational noise or redundant information.
    \item Semantic Decomposition: Splitting multi-topic or semantically complex dialogue segments into independent atomic facts, each expressing a single coherent proposition.
\end{itemize}

\subsubsection{Dataset Statistics and Examples}
The final constructed dataset $\mathcal{D}$ consists of 4,352 high-quality samples. Each sample contains a structured instruction, the dialogue context with metadata, and the target output. Figure \ref{prompt:in-out} presents a representative training instance that illustrates the model's capability to resolve context-dependent references. Specifically, the example demonstrates how the model simultaneously disambiguates pronouns (e.g., resolving "I" and "my" to "Emma") and grounds relative timestamps (e.g., converting "last Friday" into an absolute date) to produce a standalone, objective fact.
\begin{figure*}[t]
  \centering
  \begin{minipage}{0.98\textwidth}
  \begin{promptbox}[prompt:teacher_rules]{System Prompt for Atomic Fact Extraction}
Task: Extract high-value and self-contained facts from the current message, using recent conversation context to resolve contextual dependencies.

Extraction Guidelines & Rules:

1. Filter low-value content
- Ignore "noise" (greetings, fillers, acknowledgments, repetitions, conversational fluff, and questions that contain no factual information.).
- If no high-value information exists, output an empty list.

2. Ensure faithfulness
- Avoid over-inference and unsupported speculation.

3. Resolve contextual references
- Replace all pronouns (it, they, this, that) and temporal markers (yesterday, now, last week) with explicit entities and absolute dates based on the metadata or context.

3. Rewrite in third person
- Convert all first-person statements into third-person facts. Attributes must be linked to a specific name, e.g., "The user", "Alice" or "Bob".
- Include explicit attribution for subjective content, e.g., "The user believes..." or "Alice expressed frustration regarding...".
- Ensure that every fact can be understood and used independently of the original conversational context.

4. Multimodal Integration
- If [Image Description] or [Image Keywords] are present, treat the visual data as a primary source. Synthesize the visual information naturally into the text facts.

5. Output
- If multiple distinct facts exist, extract them separately.
- Output strictly in JSON format.

Now extract facts from the following conversation:
  \end{promptbox}
  \end{minipage}
  \caption{
    \textbf{System prompt for atomic fact extraction.}
    The prompt instructs the Atomic Fact Extractor to filter low-value content, resolve references, rewrite extracted information as standalone third-person facts, integrate multimodal evidence, and output the extracted facts in JSON format.
  }
  \label{fig:teacher_prompt}
\end{figure*}

\begin{figure*}[t]
  \centering
  \begin{minipage}{0.98\textwidth}
  \small
  \begin{promptbox}[prompt:in-out]{Training Sample from Dataset $\mathcal{D}$}
{
    "instruction": "Extract high-value factual information from the given dialogue and rewrite it as objective third-person facts.\n\nRequirements:\n- Ignore greetings, pleasantries, acknowledgements, and other low-value content.\n- Use dialogue context to infer implicit references, causality, or speaker intent when necessary.\n- Resolve all pronouns and vague references into explicit entities.\n- Infer specific time information when possible.\n- Each fact must be complete and standalone.\n- Output the result as a JSON array of strings.\n\nINFORMATION TO EXTRACT FROM Current Dialogue:\n[D1:1] Speaker: Emma\nTime: 1:40 pm on 17 May, 2023\nDialogue: \"Hey Zoe, guess what! I got an A on my first psychology exam last Friday! How's your week been?\"\n\nPlease extract all factual information from the current dialogue and output as a JSON array.",
    "output": "[\"Emma got an A on her first psychology exam on the Friday before May 17, 2023.\"]"
}
  \end{promptbox}
  \end{minipage}
  \caption{
    \textbf{Training sample from dataset $\mathcal{D}$.}
    The example shows an instruction-output pair used for training the Atomic Fact Extractor to extract standalone, high-value third-person facts from dialogue.
  }
  \label{fig:training_sample}
\end{figure*}

\subsection{Comparative Analysis of Fact Extraction Methods}
To validate the design of the Atomic Fact Extractor, we compare four fact extraction methods on the LoCoMo benchmark: a rule-based extractor, a zero-shot prompt extractor, a teacher-model extractor, and our proposed SFT-based extractor. All variants use the same AtomMem-Flat retrieval pipeline, with only the extraction method changed, thereby isolating the effect of the extraction module.

\begin{itemize} 
\item \textbf{Rule-based Extractor:} Filters out low-value expressions, such as greetings and acknowledgements, through phrase matching, and then splits the remaining text into sentences.
\item \textbf{Zero-shot Prompt Extractor:} Uses Qwen3-14B, the same base model as our SFT extractor, but without SFT, with the prompt in Figure \ref{prompt:in-out}.
\item \textbf{Teacher-model Extractor:} Uses GPT-4o with the teacher prompt in Figure \ref{prompt:teacher_rules}. 
\end{itemize}

\begin{table*}[t] 
\centering 
\resizebox{\textwidth}{!}{ 
\begin{tabular}{lcccccccccccccc} 
\toprule 
\multirow{2}{*}{\textbf{Method}} & \multicolumn{4}{c}{\textbf{SINGLE-HOP}} & \multicolumn{4}{c}{\textbf{MULTI-HOP}} & \multicolumn{4}{c}{\textbf{TEMPORAL}} & \multicolumn{2}{c}{\textbf{OPEN-DOMAIN}} \\ 
\cmidrule(lr){2-5} \cmidrule(lr){6-9} \cmidrule(lr){10-13} \cmidrule(lr){14-15} 
& \textbf{F1} & \textbf{BLEU-1} & \textbf{R@10} & \textbf{J} & \textbf{F1} & \textbf{BLEU-1} & \textbf{R@10} & \textbf{J} & \textbf{F1} & \textbf{BLEU-1} & \textbf{R@10} & \textbf{J} & \textbf{R@10} & \textbf{J} \\ 
\midrule 
Rule-based Extractor & 37.10 & 32.55 & 61.00 & 48.51 & 26.82 & 19.33 & 37.16 & 42.91 & 50.74 & 44.24 & 68.07 & 51.71 & 37.80 & 45.83 \\ 
Zero-shot Prompt Extractor & 44.13 & 38.09 & 69.07 & 65.16 & 31.29 & 23.76 & 41.05 & 51.06 & 54.58 & 48.04 & 75.10 & 57.63 & 37.88 & 47.92 \\ 
Teacher-model Extractor & 46.85 & 40.24 & 69.13 & 65.64 & 33.25 & 25.39 & 40.12 & 53.55 & 55.42 & 48.57 & 75.36 & 58.26 & 34.30 & 50.00 \\ 
\textbf{SFT-based Extractor} & \textbf{47.08} & \textbf{40.40} & \textbf{72.18} & \textbf{67.66} & \textbf{37.03} & \textbf{29.50} & \textbf{41.75} & \textbf{55.67} & \textbf{56.45} & \textbf{48.98} & \textbf{79.89} & \textbf{59.50} & \textbf{42.46} & \textbf{52.08} \\ 
\bottomrule 
\end{tabular} 
}
\caption{Performance comparison of four different extraction methods on the LoCoMo benchmark. R@10 denotes Recall@10. All variants use GPT-4o-mini as the answer generator and DeepSeek-V4-Pro as the evaluator.} 
\label{tab:extractor_ablation} 
\end{table*}

As shown in Table \ref{tab:extractor_ablation}, AtomMem-Flat with the SFT-based extractor achieves the highest performance across all reported metrics. In particular, it consistently outperforms the zero-shot extractor using the same base model, demonstrating the benefit of task-specific supervised fine-tuning. It also surpasses direct extraction with the GPT-4o teacher model under the same downstream retrieval pipeline. Overall, these results demonstrate the effectiveness of our SFT-based extractor. 

\subsection{Details of Event Memory Construction}
\label{event-memory}
This section details the complete algorithmic workflow for constructing and updating event memory in AtomMem as outlined in Algorithm \ref{alg:event_update}. Upon receiving a verified new atomic fact $F_{new}$, the system aims to integrate it into the broader episodic context. To achieve this, AtomMem leverages the retrieved relevant facts $\mathcal{C}_{ret}$ to track potential event associations.

\textbf{Candidate Generation.} The system forms a candidate set $\mathcal{E}_{cand}$ based on these retrieved facts. If a retrieved fact belongs to existing events, those events are added to the candidate set. If a retrieved fact is not yet linked to any event, the system includes it as a standalone candidate.

\textbf{Semantic Routing.} AtomMem prompts the LLM to verify whether $F_{new}$ logically aligns with any items in this candidate set. Based on this analysis, the LLM makes routing decisions. It can select existing events to absorb $F_{new}$ or select standalone facts to trigger the creation of new events. The model possesses the flexibility to select multiple candidates if $F_{new}$ provides different episodic contexts. Alternatively, it can choose none, opting to store the new fact independently without event affiliation.

\textbf{State Execution.} AtomMem applies these selections to update the memory state. For absorption, it appends $F_{new}$ to the member facts of all affected events. The system then prompts the LLM to regenerate the summary $\mathcal{S}$ and keywords $\mathcal{K}_e$ based on the newly added information. Concurrently, set union operations expand the participant set $\mathcal{P}_e$ and the temporal span $\mathcal{T}_e$. For new event triggers, AtomMem iterates through each matched standalone fact and instantiates a separate new event pairing it with $F_{new}$ while initializing all required event attributes.
\begin{algorithm*}[!t]
   \caption{Event Memory Construction and Update}
   \label{alg:event_update}
   \footnotesize
\begin{algorithmic}[1]
   \STATE \textbf{Input:} Verified new fact $F_{\mathrm{new}}$, retrieved context $\mathcal{C}_{\mathrm{ret}}$ containing the top-$k$ retrieved facts
   \STATE \textbf{Output:} Updated memory system state

   \STATE Initialize event candidate set $\mathcal{E}_{\mathrm{cand}} \leftarrow \emptyset$
   \STATE Initialize standalone fact set $\mathcal{F}_{\mathrm{alone}} \leftarrow \emptyset$

   \FORALL{$F \in \mathcal{C}_{\mathrm{ret}}$}
       \IF{$F$ is linked to existing events}
           \STATE $\mathcal{E}_{\mathrm{cand}} \leftarrow \mathcal{E}_{\mathrm{cand}} \cup \operatorname{GetEvents}(F)$
       \ELSE
           \STATE $\mathcal{F}_{\mathrm{alone}} \leftarrow \mathcal{F}_{\mathrm{alone}} \cup \{F\}$
       \ENDIF
   \ENDFOR

   \STATE $\mathcal{C}_{\mathrm{target}} \leftarrow \mathcal{E}_{\mathrm{cand}} \cup \mathcal{F}_{\mathrm{alone}}$

   \STATE $\mathcal{E}_{\mathrm{match}}, \mathcal{F}_{\mathrm{match}}
   \leftarrow \operatorname{LLM}_{\mathrm{verify}}(F_{\mathrm{new}}, \mathcal{C}_{\mathrm{target}})$

   \IF{$\mathcal{E}_{\mathrm{match}} \neq \emptyset$}
       \FORALL{$E \in \mathcal{E}_{\mathrm{match}}$}
           \STATE $E.\mathcal{F}_{\mathrm{ids}} \leftarrow E.\mathcal{F}_{\mathrm{ids}} \cup \{F_{\mathrm{new}}.\mathrm{id}\}$
           \STATE $E.\mathcal{P}_{e} \leftarrow E.\mathcal{P}_{e} \cup F_{\mathrm{new}}.\mathcal{P}$
           \STATE $E.\mathcal{T}_{e} \leftarrow \operatorname{TimeUnion}(E.\mathcal{T}_{e}, F_{\mathrm{new}}.\mathcal{T})$
           \STATE $E.\mathcal{S}, E.\mathcal{K}_{e}
           \leftarrow \operatorname{LLM}_{\mathrm{summarize}}(E.\mathcal{F}_{\mathrm{ids}})$
       \ENDFOR
   \ENDIF

    \IF{$\mathcal{F}_{match} \neq \emptyset$}
       \FORALL{$F_{alone} \in \mathcal{F}_{match}$}
           \STATE $\text{CreateNewEvent}(\{F_{new}, F_{alone}\})$
       \ENDFOR
   \ENDIF

   \IF{$\mathcal{E}_{\mathrm{match}} = \emptyset \land \mathcal{F}_{\mathrm{match}} = \emptyset$}
       \STATE $\operatorname{StoreIndependently}(F_{\mathrm{new}})$
   \ENDIF
\end{algorithmic}
\end{algorithm*}

\subsection{Details of Associative Graph Retrieval}
\label{sec:appendix_graph_retrieval}
Building upon the memory graph concepts introduced in Section \ref{fact_graph}, this section details the ranking of candidate facts through a multi-channel fact graph and Personalized PageRank (PPR). To execute this, AtomMem constructs a localized, query-specific subgraph anchored by the initial seed facts $\mathcal{R}_{seed}$. From these seeds, the graph expands by retrieving candidate neighbors across three relational channels: Entity, Event, and Temporal. These channels respectively capture topical similarity, event co-occurrence, and local temporal proximity.
\subsubsection{Query-Aware Keyword Weighting}
As defined in Section \ref{fact_graph}, the entity edge weight relies on a query-aware function $\omega(k)$ to balance term specificity and relevance. Simply using Term Frequency-Inverse Document Frequency (TF-IDF) can inadvertently bridge unrelated facts through high-frequency conversational stop-words. Therefore, $\omega(k)$ is implemented as a composite function:
\begin{equation} 
\omega(k) = \mathrm{IDF}(k) \cdot \beta_q(k) \cdot \pi(k) 
\end{equation}
where $\mathrm{IDF}(k)$ represents the global inverse document frequency of the keyword. $\beta_q(k)$ is a boosting multiplier that amplifies the weight if $k$ is present in the keyword set of the user query, otherwise $\beta_q(k)=1$. Furthermore, $\pi(k)$ applies a nonlinear frequency penalty that aggressively decays the weight of keywords whose occurrence frequency exceeds a certain threshold across the memory bank. This ensures that entity edges reflect genuine semantic overlap.

\subsubsection{Multi-Channel Transition Matrix}
The memory graph contains three distinct topological channels: Entity, Event, and Temporal edges. Simply adding the edge weights together would cause dense channels to overwhelm sparse ones. To address this issue, AtomMem treats them as parallel transition matrices.

For each channel $c \in \{\mathrm{ent}, \mathrm{event}, \mathrm{temp}\}$, we first independently normalize the edge weights into a channel-specific transition matrix $\mathbf{P}_c$, where each row sums to 1. The overall transition matrix $\mathbf{P}$ is then computed as a weighted fusion of the available channels:
\begin{equation} 
  \mathbf{P}_{i,j} = \sum_{c \in \mathcal{C}_i} \bar{\rho}_{i,c} (\mathbf{P}_{c})_{i,j} 
\end{equation}
where $\mathcal{C}_i$ denotes the specific set of channels containing at least one outgoing neighbor for node $F_i$. To compute the dynamic channel prior $\bar{\rho}_{i,c}$, the static prior $\rho_c$ is adaptively renormalized over these valid channels:
\begin{equation} 
\bar{\rho}_{i,c} 
= 
\frac{\rho_c}{\sum_{c' \in \mathcal{C}_i}\rho_{c'}}. 
\end{equation}
This node-wise normalization ensures that transition probabilities are properly normalized over the available outgoing channels, even when a fact does not possess all edge types.
\subsubsection{Execution of Personalized PageRank}
To initialize the PPR process, the retrieval scores of the seed facts in $\mathcal{R}_{seed}$ (obtained from the Primary and Compensatory recall stages) are transformed into a personalized restart probability distribution $\mathbf{p}$. Specifically, the system applies a power transformation to the raw seed score $\tilde{s}_i$: 
\begin{equation}
  p_i = \frac{\tilde{s}_i^{\gamma_s}}{\sum_{F_j \in \mathcal{R}_{seed}} \tilde{s}_j^{\gamma_s}}
\end{equation}
where the scaling exponent $\gamma_s$ dictates the sharpness of the distribution. This design ensures that high-confidence seeds exert a stronger restart attraction during the random walk, while still allowing the graph structure to propagate probability mass to non-seed facts that are semantically, event-wise, or temporally related.

Let $\mathbf{r}^{(t)}$ denote the vector of activation scores for all nodes in the localized graph at iteration $t$. The PageRank iteration takes the form of a Random Walk with Restart (RWR), updating the activation scores iteratively:
\begin{equation} 
\mathbf{r}^{(t+1)} = \eta \mathbf{p} + (1 - \eta) \mathbf{P}^T \mathbf{r}^{(t)} 
\end{equation}
where $\eta \in (0, 1)$ is the restart probability. A higher $\eta$ forces the walk to stay close to the seed facts, preventing the context from drifting too far from the original query intent. For dangling nodes without valid outgoing edges, their probability mass is redistributed according to the restart distribution $\mathbf{p}$, which prevents probability mass leakage during the random walk.

The iteration proceeds until it reaches a maximum number of steps or satisfies the convergence criterion:
\begin{equation} 
\|\mathbf{r}^{(t+1)} - \mathbf{r}^{(t)}\|_1 < \epsilon_c 
\end{equation}
where $\epsilon_c$ is a predefined convergence threshold. Upon convergence, the stationary distribution $\mathbf{r}^{(\infty)}$ assigns a final activation score to each fact in the local graph. AtomMem subsequently ranks all nodes according to $\mathbf{r}^{(\infty)}$ and extracts the top $k_f$ facts to construct the final retrieved context $\mathcal{R}_{fact}$. Ultimately, this framework ensures that the retrieval process remains tightly anchored to the query intent via the personalized restart distribution. Within the graph, Entity Edges drive topic-level propagation, Event Edges facilitate cross-turn episodic aggregation, and Temporal Edges preserve short-range contextual continuity. Consequently, AtomMem effectively surfaces indirectly related facts in long conversational scenarios that isolated similarity retrieval typically misses, while mitigating semantic drift through localized subgraph constraints and nonlinear frequency penalties.

\subsection{Generalizability Analysis}
\subsubsection{Cross-Model Evaluation on LoCoMo}
The main experiments use GPT-4o-mini as the backbone model for all methods. To further evaluate the generalizability of our framework, we repeat the full LoCoMo evaluation using Qwen3.5-Flash. All other experimental settings remain strictly identical to the primary evaluation detailed in Section \ref{sec:experiments}.
\begin{table*}[t]
  \centering
  \small
  \setlength{\tabcolsep}{4.2pt}
  \renewcommand{\arraystretch}{1.08}
  \resizebox{\textwidth}{!}{
  \begin{tabular}{l*{3}{ccc}c}
    \toprule
    \multirow{2}{*}{\textbf{\textsc{Method}}}
    & \multicolumn{3}{c}{\textsc{Single-Hop}}
    & \multicolumn{3}{c}{\textsc{Multi-Hop}}
    & \multicolumn{3}{c}{\textsc{Temporal}}
    & \multicolumn{1}{c}{\textsc{Open-Domain}} \\
    \cmidrule(lr){2-4}
    \cmidrule(lr){5-7}
    \cmidrule(lr){8-10}
    \cmidrule(lr){11-11}
    & $F_1 \uparrow$ & BLEU-1 $\uparrow$ & $J \uparrow$
    & $F_1 \uparrow$ & BLEU-1 $\uparrow$ & $J \uparrow$
    & $F_1 \uparrow$ & BLEU-1 $\uparrow$ & $J \uparrow$
    & $J \uparrow$ \\
    \midrule

    \textsc{LoCoMo}
    & 37.90 & 32.23 & 52.79
    & 22.82 & 15.49 & 42.20
    & 31.17 & 24.84 & 40.81
    & 25.00 \\

    \textsc{MemoryBank}
    & 21.39 & 18.21 & 30.80
    & 20.60 & 14.12 & 42.91
    & 22.55 & 19.44 & 22.12
    & 26.04 \\

    \textsc{A-Mem}
    & 41.05 & 34.41 & 58.26
    & 25.16 & 17.03 & 57.09
    & 22.49 & 16.88 & 32.40
    & 40.63 \\

    \textsc{Mem0}
    & \underline{51.04} & \underline{46.37} & \underline{70.04}
    & \underline{35.81} & 24.79 & \underline{61.35}
    & 34.55 & 32.20 & 31.46
    & 44.79 \\

    \textsc{MemoryOS}
    & 48.93 & 43.94 & 64.68
    & 35.04 & \underline{26.43} & 56.03
    & 29.83 & 25.01 & 33.64
    & 46.88 \\

    \textsc{LightMem}
    & 41.43 & 35.89 & 60.05
    & 26.03 & 21.34 & 48.23
    & 45.40 & 40.83 & 56.70
    & 40.63 \\

    \midrule
    \textsc{AtomMem-Flat}
    & 49.45 & 42.13 & 67.66
    & 35.32 & 25.50 & 58.51
    & \underline{50.81} & \underline{42.81} & \underline{61.99}
    & \underline{48.96} \\

    \textsc{AtomMem}
    & \textbf{55.31} & \textbf{49.05} & \textbf{73.01}
    & \textbf{38.17} & \textbf{30.91} & \textbf{62.06}
    & \textbf{55.97} & \textbf{51.60} & \textbf{63.55}
    & \textbf{58.33} \\

    \bottomrule
  \end{tabular}
  }
  \caption{
    Performance comparison on LoCoMo using Qwen3.5-Flash as the backbone model and DeepSeek-V4-Pro as the evaluator. The best and second-best results are highlighted in bold and underlined.
  }
  \label{tab:qwen35_locomo}
\end{table*}

Table \ref{tab:qwen35_locomo} presents the cross-model evaluation results on LoCoMo using Qwen3.5-Flash as the backbone model. AtomMem consistently achieves the best overall performance across different question types, showing that the observed gains are not merely a result of backbone-specific LLM preferences, but stem from the architectural advantages of AtomMem. Its atomic facts consistently provide high-quality memory representations across different backbone models, leading to performance improvements. Overall, these strong results further highlight the generalizability and robustness of our framework.

\subsubsection{Cross-Benchmark Evaluation on LongMemEval}
\begin{table*}[!t]
  \centering
  \small
  \setlength{\tabcolsep}{10pt}
  \renewcommand{\arraystretch}{1.1}
  \begin{tabular*}{0.86\textwidth}{@{\extracolsep{\fill}}lrrr@{}}
    \toprule
    \textbf{Stage} & \textbf{Avg (ms)} & \textbf{P95 (ms)} & \textbf{Max (ms)} \\
    \midrule
    Total online latency & 3585.11 & 4559.80 & 9596.74 \\
    \midrule
    Query intent & 2079.86 & 2690.29 & 5399.86 \\
    Answer generation & 1346.14 & 1970.20 & 7678.18 \\
    \midrule
    Retrieval pipeline only & 145.80 & 235.87 & 3107.25 \\
    \quad Query embedding & 13.23 & 20.71 & 1018.54 \\
    \quad Base retrieval & 35.29 & 89.54 & 2933.75 \\
    \quad Graph rerank & 109.90 & 184.16 & 2002.02 \\
    \bottomrule
  \end{tabular*}
  \caption{
    End-to-end latency breakdown of the AtomMem system across different processing stages.
  }
  \label{tab:latency_breakdown}
\end{table*}
\label{longmemeval}
To demonstrate that our system does not suffer from data bias toward the LoCoMo benchmark, we conduct supplementary evaluations on the completely independent LongMemEval dataset. We compare AtomMem with two strong baselines, LightMem and Mem0, under the same evaluation setting, using GPT-4o-mini as the backbone model and DeepSeek-V4-Pro as the judge model. Notably, the SSP category evaluates subjective and personalized generation quality, making traditional token-matching metrics such as the $F_1$ score and BLEU-1 inapplicable. We therefore report only the LLM-as-a-Judge score $J$ for this category.

Table \ref{tab:longmemeval_comparison} presents the detailed performance across the six distinct question categories defined in the LongMemEval framework. AtomMem consistently outperforms both strong baselines across all categories. These results provide further evidence that AtomMem generalizes beyond LoCoMo and remains effective across diverse long-term memory tasks.

\begin{table*}[t]
  \centering
  \small
  \setlength{\tabcolsep}{5.2pt}
  \renewcommand{\arraystretch}{1.08}
  \resizebox{\textwidth}{!}{
  \begin{tabular}{lccc ccc ccc}
    \toprule
    \multirow{2}{*}{\textbf{\textsc{Category}}}
    & \multicolumn{3}{c}{\textbf{\textsc{AtomMem}}}
    & \multicolumn{3}{c}{\textsc{LightMem}}
    & \multicolumn{3}{c}{\textsc{Mem0}} \\
    \cmidrule(lr){2-4}
    \cmidrule(lr){5-7}
    \cmidrule(lr){8-10}
    & $F_1 \uparrow$ & BLEU-1 $\uparrow$ & $J \uparrow$
    & $F_1 \uparrow$ & BLEU-1 $\uparrow$ & $J \uparrow$
    & $F_1 \uparrow$ & BLEU-1 $\uparrow$ & $J \uparrow$ \\
    \midrule
    SSU & \textbf{80.70} & \textbf{75.84} & \textbf{90.00}
        & 20.97 & 12.67 & 84.29
        & 21.95 & 13.61 & \textbf{90.00} \\
    MS  & \textbf{57.50} & \textbf{54.85} & \textbf{67.67}
        & 6.72 & 4.22 & 57.14
        & 6.68 & 4.60 & 65.41 \\
    TR  & \textbf{42.10} & \textbf{33.16} & \textbf{52.63}
        & 12.74 & 7.65 & 51.88
        & 8.55 & 5.45 & 48.12 \\
    KU  & \textbf{66.35} & \textbf{62.27} & \textbf{79.49}
        & 13.34 & 8.21 & 76.92
        & 12.57 & 7.78 & 74.36 \\
    SSA & \textbf{61.09} & \textbf{52.17} & \textbf{76.79}
        & 9.14 & 5.56 & 33.93
        & 10.32 & 6.65 & 44.64 \\
    SSP & -- & -- & \textbf{80.00}
        & -- & -- & 60.00
        & -- & -- & 63.33 \\
    \bottomrule
  \end{tabular}
  }
  \caption{ Performance comparison of AtomMem against two strong baselines on the LongMemEval benchmark. The best results are highlighted in bold. The abbreviations correspond to Single-Session-User (SSU), Multi-Session (MS), Temporal-Reasoning (TR), Knowledge-Update (KU), Single-Session-Assistant (SSA), and Single-Session-Preference (SSP).
  }
   \label{tab:longmemeval_comparison}
\end{table*}

\subsection{Update Stability Analysis}
To evaluate the stability of our memory update mechanism, we conduct a controlled ablation on update strategies. We compare our proposed Fact Verification mechanism against a broader Same-Item Update baseline. The Same-Item Update strategy rewrites an existing fact when new information concerns the same item or event and changes or extends its content. In contrast, Fact Verification updates an existing fact only when a logical conflict is detected. All variants build upon the AtomMem-Flat architecture and process the same pre-extracted LoCoMo facts, thereby isolating the effect of the update strategy.

As shown in Table~\ref{tab:update_stability}, Same-Item Update performs 959 rewrites, more than three times the 301 updates performed by Fact Verification. Despite its higher update frequency, Same-Item Update underperforms Fact Verification across all reported metrics and performs worse than AtomMem-Flat on most metrics. In contrast, Fact Verification achieves the best performance across all question types while modifying fewer existing facts. These results indicate that broader and more frequent rewriting does not necessarily improve memory quality. Our conflict-driven update mechanism limits unnecessary modifications while preserving useful information, thereby reducing the risk of update-induced semantic drift and maintaining stronger downstream performance.
\begin{table*}[t]
  \centering
  \small
  \setlength{\tabcolsep}{3.2pt}
  \renewcommand{\arraystretch}{1.08}
  \resizebox{\textwidth}{!}{
  \begin{tabular}{l r *{3}{cccc} cc}
    \toprule
    \multirow{2}{*}{\textbf{\textsc{Method}}}
    & \multirow{2}{*}{\textbf{\shortstack{\textsc{Update}\\\textsc{Count}}}}
    & \multicolumn{4}{c}{\textsc{Single Hop}}
    & \multicolumn{4}{c}{\textsc{Multi-Hop}}
    & \multicolumn{4}{c}{\textsc{Temporal}}
    & \multicolumn{2}{c}{\textsc{Open Domain}} \\
    \cmidrule(lr){3-6}
    \cmidrule(lr){7-10}
    \cmidrule(lr){11-14}
    \cmidrule(lr){15-16}
    & & $F_1$ & BLEU-1 & R@10 & $J$
      & $F_1$ & BLEU-1 & R@10 & $J$
      & $F_1$ & BLEU-1 & R@10 & $J$
      & R@10 & $J$ \\
    \midrule

    \textsc{AtomMem-Flat}
    & 0
    & 47.08 & 40.40 & 72.18 & 67.66
    & 37.03 & 29.50 & 41.75 & 55.67
    & 56.45 & 48.98 & 79.89 & 59.50
    & 42.46 & 52.08 \\

    \quad + Same-Item Update
    & 959
    & 47.75 & 41.39 & 70.27 & 66.11
    & 35.61 & 27.96 & 42.31 & 54.26
    & 53.93 & 47.77 & 78.43 & 56.70
    & 39.07 & 50.00 \\

    \quad + Fact Verification
    & 301
    & \textbf{49.08} & \textbf{42.85} & \textbf{73.24} & \textbf{68.37}
    & \textbf{37.76} & \textbf{29.66} & \textbf{43.58} & \textbf{56.38}
    & \textbf{57.59} & \textbf{50.78} & \textbf{80.20} & \textbf{62.30}
    & \textbf{43.16} & \textbf{53.13} \\

    \bottomrule
  \end{tabular}
  }
  \caption{
    Controlled comparison of memory update strategies on LoCoMo.
    All variants use the same pre-extracted atomic facts and flat embedding-based retrieval pipeline.
    Update Count denotes the number of modifications made to existing facts.
    R@10 and $J$ denote Recall@10 and LLM-as-a-Judge, respectively.
    The best results are highlighted in bold.
  }
  \label{tab:update_stability}
\end{table*}

\subsection{End-to-End Efficiency Analysis}
\label{sec:efficiency}
To evaluate the practical deployment viability of AtomMem, we conduct a comprehensive latency analysis across all processing stages using 1,540 questions from the LoCoMo dataset. The experiments are executed locally on an Intel Core i7-10750H CPU (6 physical cores, 12 logical threads) to accurately evaluate the computational overhead.

As detailed in Table \ref{tab:latency_breakdown}, the total online processing latency averages approximately 3.6 seconds. The vast majority of this execution time is consumed by the fundamental LLM inference calls. In stark contrast, our core memory retrieval pipeline introduces negligible computational overhead. The entire retrieval process averages only 145.80 milliseconds. Furthermore, our graph-based reranking mechanism incurs limited computational overhead, requiring only 109.90 milliseconds on average. These results indicate that AtomMem maintains practical retrieval efficiency despite its additional structural components, making it suitable for real-time interactive applications.

\subsection{Hyperparameter Sensitivity Analysis}
\label{sec:hyperparameter_analysis}
Complementing the hyperparameter analysis provided in the main text, this section presents supplementary evaluations to determine the optimal configurations for other critical components within the AtomMem retrieval pipeline. Specifically, we investigate the sensitivity of the overall system performance to the initial seed count for local graph construction and the compensatory fusion weight. All experiments are evaluated on the LoCoMo benchmark.
\subsubsection{Impact of Graph Retrieval Initial Seed Count $k_s$}
\label{sec:seed_optimization}
To investigate the impact of the initial seed count on local graph construction, we conduct a detailed parametric sweep over the total seed capacity $k_s$. This hyperparameter directly controls the number of entry points injected into the personalized PageRank algorithm. 

\begin{figure}[t]
  \centering
  \includegraphics[width=0.85\linewidth]{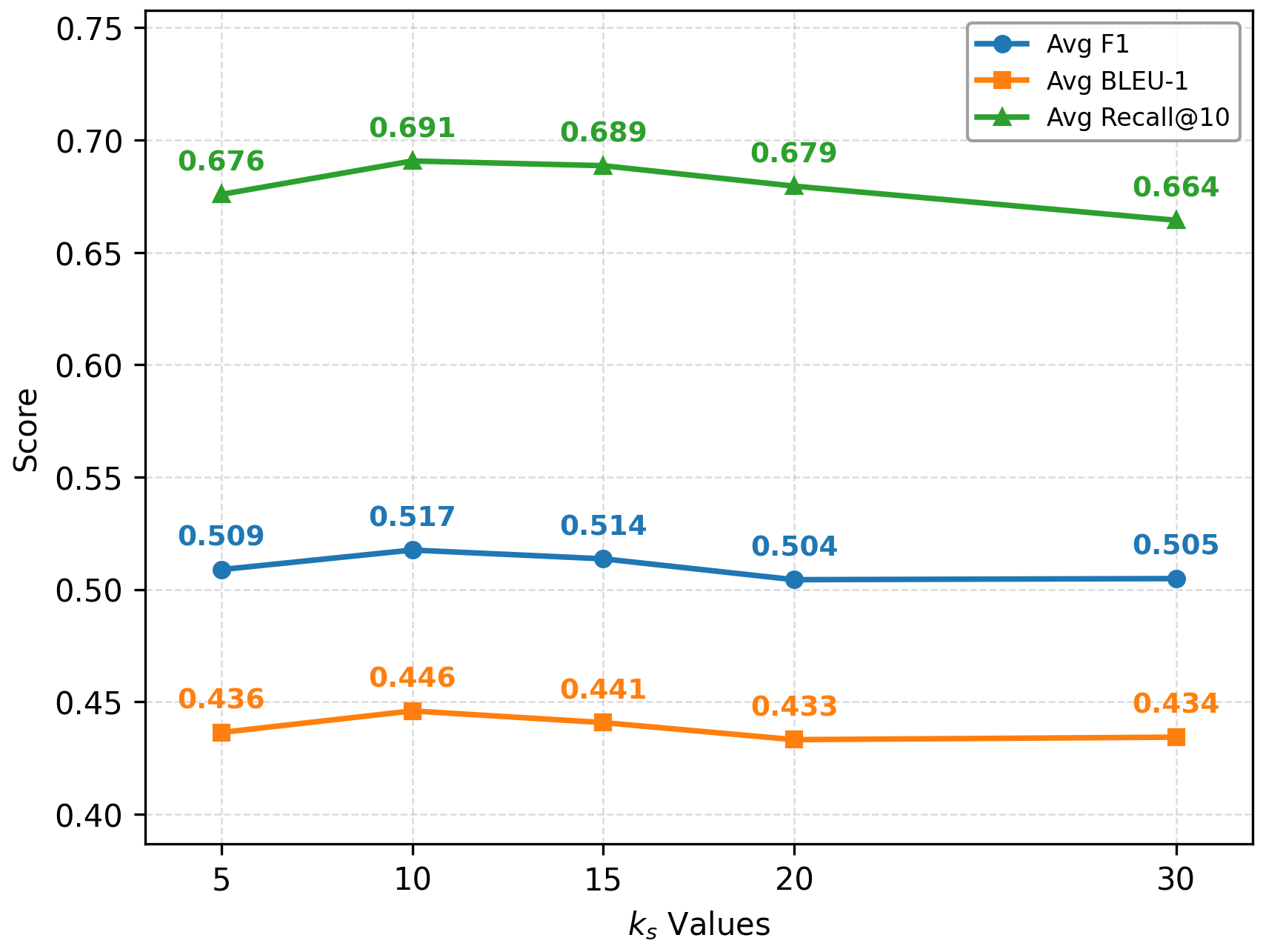}
  \caption{
    Hyperparameter analysis of the graph retrieval initial seed count $k_s$. The performance curves indicate that $k_s=10$ achieves the optimal trade-off between informational coverage and noise suppression.
  }
  \label{fig:seed_optimization}
\end{figure}

As illustrated in Figure \ref{fig:seed_optimization}, the system performance varies systematically with respect to the seed density. When $k_s$ is set to a conservative value of 5, the initial memory activation remains insufficient, leading to suboptimal recall metrics. Conversely, expanding the seed count beyond 10 introduces a performance degradation across all evaluation dimensions, where the aggregate semantic accuracy metrics such as average F1 and BLEU-1 experience notable declines. This trend demonstrates that excessive initial seeds inevitably pollute the local graph structure with irrelevant conversational contexts and redundant relational edges. This semantic noise degrades the random walk propagation, allowing unrelated facts to accumulate high stationary probabilities. Consequently, optimizing the configuration at $k_s = 10$, where the main seed and compensation seed are both set to 5, provides the ideal trade-off between informational coverage and noise suppression. We therefore freeze $k_s = 10$ as the default setting for all primary experiments.   

\subsubsection{Impact of Compensatory Fusion Weight $w_e$}
To systematically determine the optimal balance between global event relevance and local fact precision during the compensatory recall phase, we analyze the sensitivity of the retrieval performance to the fusion weight $w_e$. This hyperparameter controls the proportional contribution of the event-level hybrid score and the fact-level hybrid score when ranking candidate facts extracted from matched events.

\begin{figure}[t]
  \centering
  \includegraphics[width=0.85\linewidth]{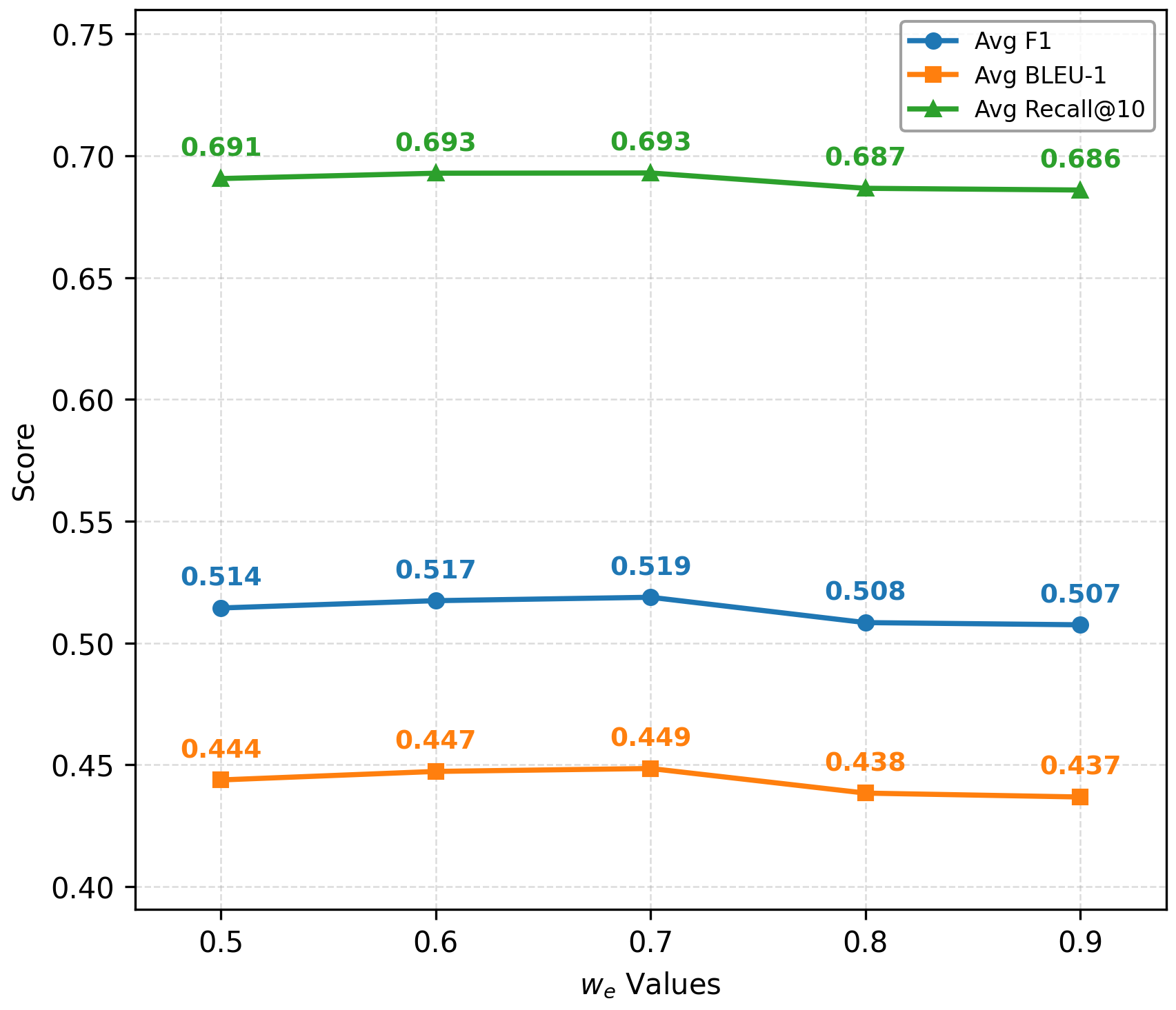}
  \caption{
    Hyperparameter analysis of the compensatory fusion weight $w_e$. The performance curves indicate that $w_e=0.7$ achieves the optimal balance between global event relevance and local fact precision.
  }
  \label{fig:fusion_weight}
\end{figure}

As illustrated in Figure \ref{fig:fusion_weight}, the system performance varies systematically across different $w_e$ configurations. The metrics reveal a steady improvement as $w_e$ increases from 0.5 to 0.7, reaching a distinct optimal peak at $w_e = 0.7$ across all core evaluation dimensions, including average F1, BLEU-1, and Recall@10. Conversely, elevating the weight beyond 0.7 introduces a noticeable performance degradation. This trend effectively validates our hierarchical retrieval design. Specifically, when $w_e$ is set too low, the system underutilizes the broader episodic context provided by the event layer, causing the compensatory recall to regress toward a standard fact-level semantic search. When $w_e$ exceeds 0.7, the fusion score becomes overly dominated by global event relevance, forcing the system to retrieve constituent facts that belong to highly relevant events but lack direct semantic alignment with the specific user query. Consequently, optimizing the configuration at $w_e = 0.7$ provides the ideal trade-off, ensuring that the retrieved facts are both contextually grounded in the correct episode and individually precise.

\subsection{Hyperparameter Settings}
\label{app:hyperparameters}
This section details the comprehensive hyperparameter configurations used for the AtomMem system. As discussed in Section~\ref{ex_set}, while the final fact retrieval count is standardized at $k_f=10$, our hierarchical architecture requires additional parameters to govern event retrieval, profile matching, and graph-based score fusion. These specific values were empirically determined through extensive hyperparameter analysis to optimize the overall performance of the memory retrieval pipeline. To ensure full reproducibility, all configurations governing the local graph construction and random walk mechanics are explicitly detailed in Table \ref{tab:hyperparameters}.

\begin{table*}[!t]
  \centering
  \small
  \renewcommand{\arraystretch}{1.2}
  \begin{tabular*}{\textwidth}{@{\extracolsep{\fill}}llc}
    \toprule
    \textbf{Module Component} & \textbf{Hyperparameter} & \textbf{Value} \\
    \midrule
    \textbf{Base Capacities} & Initial Graph Seed Count ($k_s$) & 10 \\
    & Final Retrieved Fact Count ($k_f$) & 10 \\
    & Profile Item Retrieval Count ($k_p$) & 5 \\
    \midrule
    \textbf{Similarity Weighting} & Embedding Similarity Weight ($\alpha$) & 0.7 \\
    & Keyword Jaccard Weight ($\beta$) & 0.3 \\
    \midrule
    \textbf{Compensatory Fusion} & Event Relevance Weight ($w_{e}$) & 0.7 \\
    & Fact Self-Relevance Weight ($w_{f}$) & 0.3 \\
    \midrule
    \textbf{Graph Channel Mixture} & Entity Edge Weight ($\rho_{ent}$) & 0.45 \\
    & Event Edge Weight ($\rho_{event}$) & 0.40 \\
    & Temporal Edge Weight ($\rho_{temp}$) & 0.15 \\
    \midrule
    \textbf{Random Walk with Restart} & Restart Probability ($\eta$) & 0.34 \\
    & Maximum Iterations & 20 \\
    & Convergence Tolerance ($\epsilon_c$) & $10^{-6}$ \\
    & Seed Score Sharpening Power ($\gamma_s$) & 5.0 \\
    \midrule
    \textbf{Graph Construction} & Maximum Seed Facts & 10 \\
    & Maximum Expansion Hops & 2 \\
    & Maximum Local Graph Nodes & 180 \\
    & Maximum Neighbors per Fact & 30 \\
    & Edge Weight Epsilon & $10^{-8}$ \\
    \midrule
    \textbf{Entity Edge Weighting} & Query Keyword Boost & 2.5 \\
    & Query Keyword Penalty Floor & 0.45 \\
    & Query Keyword Penalty Threshold $\tau_q$ & 0.05 \\
    & Query Keyword Penalty Exponent $\gamma_q$ & 0.7 \\
    & Non-query Keyword Penalty Threshold $\tau_{nq}$ & 0.10 \\
    & Non-query Keyword Penalty Exponent $\gamma_{nq}$ & 1.0 \\
    \midrule
    \textbf{Event Edge Constraints} & Event Size Penalty Exponent ($\gamma_e$) & 1.25 \\
    & Maximum Event Size & 60 \\
    & Maximum Event-edge Expansion Hops & 2 \\
    \midrule
    \textbf{Temporal Constraints} & Turn Window Size ($W$) & 2 \\
    & Turn Distance Decay Temperature ($\tau$) & 2.0 \\
    & Maximum Temporal-edge Expansion Hops & 1 \\
    \bottomrule
  \end{tabular*}
  \caption{
    Comprehensive empirical hyperparameter configurations optimized for the AtomMem retrieval pipeline.
  }
  \label{tab:hyperparameters}
\end{table*}

\subsection{Prompts}
This section details the system prompts utilized throughout our experiments, covering both the response generation phase and the LLM judgment phase. 

Figure \ref{fig:prompt_generation_exact} presents the generation prompt applied to exact extraction tasks including single-hop, multi-hop, and temporal reasoning. For open-domain queries that necessitate broader reasoning and integration with external world knowledge, we utilize the prompt detailed in Figure \ref{fig:prompt_generation_open}. Additionally, Figure \ref{fig:prompt_eval} outlines the evaluation prompt used for the LLM judge to score the generated responses against the ground-truth answers.
\begin{figure*}[t]
  \centering
  \begin{minipage}{0.98\textwidth}
  \begin{promptbox}[prompt:response_generation_exact]{System Prompt for Response Generation - Single-Hop / Multi-Hop / Temporal Reasoning}
Task: Generate an answer based on retrieved information (Profiles and/or Facts).

Input:
- query: Original user query
- profiles: List of Profile statements (optional)
- facts: List of Fact objects

Your task is to answer questions in an extremely concise manner.

Please only provide the content of the answer, without including "answer:".
   
For questions that require a date or time to be answered, please strictly follow the format of "July 15, 2023" and provide specific dates as much as possible. Do not answer relative times such as 'last year' or 'last week'. Please infer the specific year/date based on the provided information, rather than just saying 'last year'. Only provide one year, date, or time, no additional answers required.

If the question is about the duration, answer in the form of several years, months, or days.

When the question is: "What did the charity race raise awareness for?", you should not answer in the form of: "The charity race raised awareness for mental health." Instead, it should be: "mental health", as this is more concise.

Now answer the following query based on the retrieved information:
\end{promptbox}
  \end{minipage}
  \caption{
    \textbf{System Prompt for Response Generation.} This prompt is utilized for single-hop, multi-hop, and temporal reasoning tasks where precise extraction is required.
  }
  \label{fig:prompt_generation_exact}
\end{figure*}

\begin{figure*}[t]
  \centering
  \begin{minipage}{0.98\textwidth}
\begin{promptbox}[prompt:response_generation_open]{System Prompt for Response Generation - Open-Domain}
Task: Generate an answer based on retrieved information (Profiles and/or Facts).

Input:
- query: Original user query
- profiles: List of Profile statements (optional)
- facts: List of Fact objects

IMPORTANT RULES:
1. Base your answer on the retrieved information:
- Use Profiles to understand: personality, preferences, habits, characteristics
- Use Facts to understand: specific behaviors, events, statements
- Combine both to infer/predict the answer when needed
- Not all information may be needed, and you must judge which parts are sufficient to answer the question.

2. REASONING WITH EXTERNAL KNOWLEDGE:
- Combine the provided information with external knowledge (such as common sense or world facts).
- Consider cause-and-effect relationships and logical implications.
- Use your understanding of human behavior, social norms, and general knowledge to enhance your answer.

OUTPUT FORMAT:
Provide a direct natural language answer without any JSON structure or additional formatting.

Now answer the following query based on the retrieved information:
\end{promptbox}
  \end{minipage}
  \caption{
    \textbf{System Prompt for Response Generation (Open-Domain).} This prompt guides the model to integrate retrieved memory with external knowledge for comprehensive reasoning.
  }
  \label{fig:prompt_generation_open}
\end{figure*}

\begin{figure*}[t]
  \centering
  \begin{minipage}{0.98\textwidth}
\begin{promptbox}[prompt:eval_answer]{System Prompt for Answer Judgment}
Your task is to label an answer to a question as "CORRECT" or "WRONG". You will be given the following data: (1) a question (posed by one user to another user), (2) a 'gold' (ground truth) answer, (3) a generated answer which you will score as CORRECT/WRONG. 

The point of the question is to ask about something one user should know about the other user based on their prior conversations.

The gold answer will usually be a concise and short answer that includes the referenced topic, for example:
Question: Do you remember what I got the last time I went to Hawaii? 
Gold answer: A shell necklace 

The generated answer might be much longer, but you should be generous with your grading - as long as it touches on the same topic as the gold answer, it should be counted as CORRECT. 

For time-related questions, the gold answer will be a specific date, month, or year. The generated answer might include relative references (e.g., "last Tuesday"), but you should be generous - if it refers to the same time period as the gold answer, mark it CORRECT, even if the format differs (e.g., "May 7th" vs. "7 May").

Now it's time for the real question: 
Question: {question} 
Gold answer: {gold_answer} 
Generated answer: {generated_answer} 

First, provide a short (one sentence) explanation of your reasoning, then finish with CORRECT or WRONG. Do NOT include both CORRECT and WRONG in your response, or it will break the evaluation script. 

Just return the label CORRECT or WRONG in a json format with the key as "label".
\end{promptbox}
  \end{minipage}
  \caption{
    \textbf{System Prompt for Answer Judgment.} This prompt configures the LLM judge to evaluate generated answers against ground-truth references. 
   }
  \label{fig:prompt_eval}
\end{figure*}
\end{document}